\documentclass{article} 
\usepackage[preprint]{colm2025_conference}  
\usepackage{algorithmicx}
\usepackage[most]{tcolorbox}
\usepackage{microtype}
\usepackage{hyperref}
\usepackage{url}
\usepackage{amsmath}
\usepackage{booktabs}
\usepackage{fontawesome}
\usepackage{graphicx} 
\usepackage{multirow} 
\usepackage{subfig}
\usepackage{colortbl}
\usepackage{color}
\usepackage{microtype}
\usepackage{lineno}
\usepackage{algorithm}
\usepackage{algpseudocode}

\definecolor{darkblue}{rgb}{0, 0, 0.5}
\hypersetup{colorlinks=true, citecolor=darkblue, linkcolor=darkblue, urlcolor=darkblue}

\title{Commander-GPT: Dividing and Routing for Multimodal Sarcasm Detection}

\begin{document}
\author{Yazhou Zhang  \\
College of Intelligence and Computing\\
Tianjin University\\
Tianjin, 300350, China \\
\texttt{yzhou\_zhang@tju.edu.cn} \\
\And
Chunwang Zou  \\
Software Engineering College \\
Zhengzhou University of Light Industry \\
Zhengzhou , 450002, China \\
\texttt{1376132667@qq.com} \\
\And
Bo Wang \\
College of Intelligence and Computing \\
Tianjin University\\
Tianjin, 300350, China \\
\texttt{bo\_wang@tju.edu.cn} \\
\And
Jing Qin  \\
Department of Computational Neuroscience \\
The Hong Kong Polytechnic University \\
Hong Kong,999077, China \\
\texttt{harry.qin@polyu.edu.hk} \\
\And
Prayag Tiwari \\
School of Information Technology \\
Halmstad University \\
\texttt{prayag.tiwari@ieee.org}
}
%

\ifcolmsubmission
\linenumbers
\fi

\maketitle

\begin{abstract}
Multimodal sarcasm understanding is a high-order cognitive task. Although large language models (LLMs) have shown impressive performance on many downstream NLP tasks, growing evidence suggests that they struggle with sarcasm understanding. 
In this paper, we propose Commander-GPT, a modular decision routing framework inspired by military command theory. Rather than relying on a single LLM's capability, Commander-GPT orchestrates a team of specialized LLM agents where each agent will be selectively assigned to a focused sub-task such as keyword extraction, sentiment analysis, etc. Their outputs are then routed back to the commander, which integrates the information and performs the final sarcasm judgment.
To coordinate these agents, we introduce three types of centralized commanders: 
(1) a trained lightweight encoder-based commander (e.g., multi-modal BERT); (2) four small autoregressive language models, serving as moderately capable commanders (e.g., DeepSeek-VL); (3) a large LLM-based commander (GPT-4o) that performs task routing, output aggregation, and sarcasm decision-making in a zero-shot fashion.
We evaluate Commander-GPT on the MMSD and MMSD 2.0 benchmarks, comparing six prompting strategies and one ensemble learning approach. Experimental results show that our framework achieves a 19.3\% improvement in F1 score over state-of-the-art (SoTA) baselines, demonstrating its effectiveness.
\end{abstract}

\section{Introduction}
The era of large language models (LLMs) has been propelled by the scaling laws of language models and the emergence of capabilities with increasing model scale. SoTA LLMs, such as GPT-4o~\citep{achiam2023gpt}, Claude 4\footnote{https://www.anthropic.com/news/claude-4}, DeepSeek R1~\citep{deepseekai2025deepseekr1incentivizingreasoningcapability}, Qwen 3~\citep{yang2025qwen3technicalreport}, et., have demonstrated remarkable performance across a wide range of downstream natural language processing (NLP) tasks, including question answering, machine translation, commonsense reasoning, and code generation. These models exhibit impressive zero-shot and few-shot generalization abilities, leading to the belief that LLMs may have reached a critical threshold of general intelligence~\citep{zhang2025mlmstrappedvisualroom}.

Despite these advances, sarcasm understanding remains a significant and unresolved challenge. Sarcasm is a nuanced linguistic phenomenon that often employs rhetorical devices such as irony, hyperbole, and contradiction to express sentiments that diverge sharply from the literal meanings of words~\citep{liu2023quantum}. For example, the sentence ``Oh great, another meeting that could have been an email.'' appears to express enthusiasm, but in reality, conveys frustration or annoyance, its sarcastic tone hinging on pragmatic context and emotional subtext. Accurate sarcasm detection requires a combination of contextual reasoning, emotional inference, and figurative language interpretation, multi-modal interaction, etc., all of which demand high-order cognitive capabilities. 
 \begin{figure*}[t]
\centering
\includegraphics[width=5.3in]{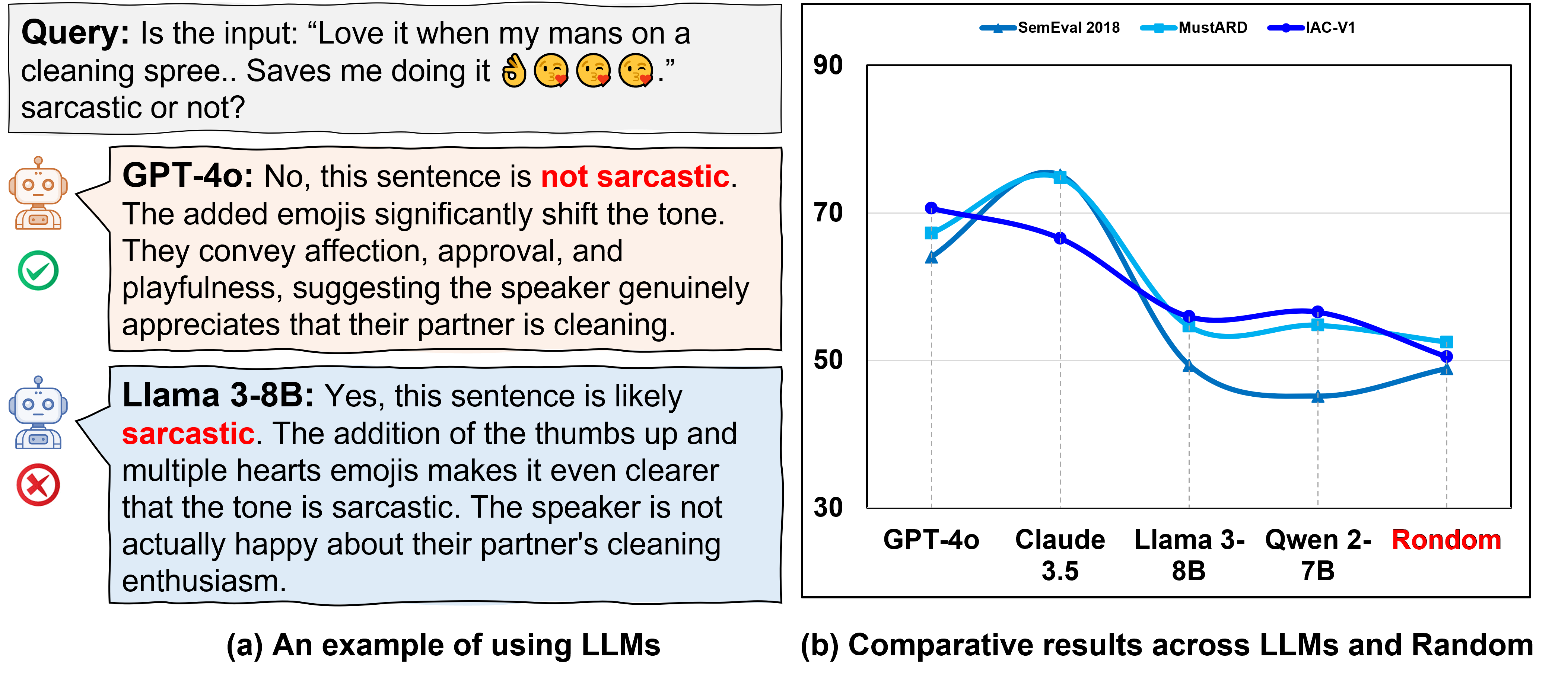}
\caption{LLM Performance on three sarcasm datasets in prior work.} \label{fig:example}
\end{figure*}

The growing evidence proves that even state-of-the-art models such as GPT-4, Claude 3.5 often perform poorly on sarcasm detection, sometimes close to random guessing, as shown in Fig.~\ref{fig:example}~\citep{yao2025sarcasm}. This observation challenges the prevailing assumption that general-purpose LLMs, by virtue of their scale, can seamlessly handle high-level pragmatic tasks. This discrepancy reveals a fundamental limitation: current methods treat sarcasm detection as an undivided, end-to-end task, relying on a single model to implicitly execute multiple layers of reasoning. These include sentiment recognition, rhetorical structure identification, contextual interpretation, and visual-textual alignment. Without explicit modeling of these sub-tasks, even powerful LLMs often fail to capture the complex and composite nature of sarcasm, particularly in multimodal scenarios.

This observation motivates a shift from monolithic processing to a modular strategy. Drawing inspiration from military command structures where ``the right person for the right job'' principle ensures mission success, we recognize that different LLMs possess distinct strengths. For example, some excel at linguistic analysis, others at visual understanding (e.g., BLIP), and still others at cross-modal reasoning (e.g., GPT-4o). Just as military commanders strategically deploy specialized units based on their unique capabilities.

To this end, we propose Commander-GPT, a structured multi-agent framework that decomposes sarcasm detection into six cognitively meaningful sub-tasks. They are: \textit{context modeling}, \textit{sentiment analysis}, \textit{rhetorical device recognition}, \textit{facial expression recognition}, \textit{image summarization}, and \textit{scene text recognition}.
Each sub-task is handled by an expert LLM or MLLM agent. Rather than invoking all agents for every input, the commander first analyzes the input and then activates only the agents that are most suitable for handling the relevant sub-tasks by introducing the routing scorer. In addition, BLIP-2~\citep{li2022blip}, Vision Transformer\footnote{\url{https://huggingface.co/motheecreator/vit-Facial-Expression-Recognition}}, and OCR-2.0~\citep{wei2024general} are selected as vision agents for image summarization, facial expression recognition, and scene text recognition, respectively, while Llama 3-8B\footnote{\url{https://huggingface.co/meta-llama/Meta-Llama-3-8B}}, GLM-2B\footnote{\url{https://huggingface.co/THUDM/glm-2b}}, and RoBERTa\footnote{\url{https://huggingface.co/SamLowe/roberta-base-go_emotions}} serve as linguistic specialists. Their outputs are then routed back to the commander, which integrates the information and performs the final sarcasm judgment.

To coordinate these agents, we introduce three types of centralized commanders: 
(1) a trained lightweight encoder-based commander (e.g., multi-modal BERT); (2) four small autoregressive language models, serving as moderately capable commanders (e.g., DeepSeek-VL-7B); (3) two large LLM-based commander (Gemini Pro and GPT-4o) that performs task routing, output aggregation, and sarcasm decision-making.
This diverse commander configuration enables a comprehensive exploration of the trade-offs between performance and scalability.

Finally, we conduct empirical evaluations of Commander-GPT on two benchmark datasets: MMSD and MMSD 2.0. We compare our framework against five state-of-the-art (SoTA) prompting strategies (e.g., Chain-of-Thought, Plan-and-Solve, $S^3$ Agent, etc.). Experimental results highlight three key observations: (1) Commander-GPT achieves 4.4\% and 11.7\% improvement in F1 score over strong baselines, demonstrating the effectiveness; (2) despite advances in large multimodal language models (MLLMs), fine-tuned small models (such as BERT) still exhibit stronger sarcasm detection capabilities; (3) the Commander-GPT model demonstrates robust generalization across diverse backbone LLMs and domains.

Our contributions are summarized as follows: 
\begin{itemize}
    \item We propose \textbf{Commander-GPT}, a modular multi-agent framework for multimodal sarcasm detection.
    \item We introduce a set of centralized commanders with varying model capacities and conduct a systematic comparison of their orchestration ability in sarcasm detection.
    \item We present extensive experiments on MMSD and MMSD 2.0, showing that Commander-GPT achieves 4.4\% and 11.7\% F1 improvement over SoTA baselines on average.
\end{itemize}

\section{Related Work}
\subsection{Multimodal Large Language Models}

Multimodal large language models (MLLMs) aim to unify language and vision understanding through joint pretraining and alignment across modalities. Early frameworks such as CLIP~\citep{radford2021learning} and BLIP~\citep{li2022blip} align vision and text representations through contrastive learning or caption supervision. These models laid the groundwork for subsequent generative MLLMs like Flamingo~\citep{alayrac2022flamingo}, MiniGPT-4~\citep{zhu2023minigpt}, and LLaVA~\citep{liu2023visual}, which integrate vision encoders with pretrained LLMs via lightweight adapters or projection layers. These models support open-ended visual question answering and caption generation.
Qwen-VL~\citep{bai2023qwen}, InternVL~\citep{chen2024internvl}, and Emu2~\citep{sun2024generative} further explore multi-granular alignment, spatial grounding, and tool-augmented multimodal reasoning. 

While MLLMs such as GPT-4o and Gemini have achieved strong performance in visual grounding and factual question answering, they consistently underperform on high-level pragmatic tasks like sarcasm and irony~\citep{yao2025sarcasm,yang2024emollm}. Sarcastic expressions often rely on rhetorical contradiction and contextual incongruity, which current models struggle to capture.
Notably, ~\cite{yao2025sarcasm} proposed advanced prompting strategies, i.e., \textit{graph-of-cue} and \textit{bagging-of-cue} to improve sarcasm detection, but found them less effective than standard input–output approach. This suggests that increasing prompt complexity alone cannot overcome the reasoning limitations of monolithic models. In contrast, our proposed Commander-GPT framework explicitly decomposes sarcasm detection into specialized sub-tasks, each handled by expert agents under centralized coordination, enabling more effective and interpretable multimodal sarcasm understanding.

\subsection{Multimodal Sarcasm Detection}

Multimodal sarcasm detection is a complex task that requires the integration of linguistic, visual, and contextual information to resolve non-literal intent and pragmatic ambiguity. In recent years, researchers have explored a variety of neural architectures and multimodal fusion strategies to address the inherent challenges of this task. 

Recent work has shifted toward more advanced architectures that better capture the nuanced interplay between modalities. For example, ~\cite{wang2024s3} proposed the \( S^3 \) Agent framework, which employs visual large language models and integrates multi-perspective analysis to enhance zero-shot multimodal sarcasm detection. Similarly, ~\cite{tang2024leveraging} utilized generative MLLMs equipped with instruction templates and demonstration retrieval to improve the understanding of complex sarcastic cues. Aggarwal et al.~\citep{aggarwal2024modelling} designed a framework capable of processing multimodal input triples—including text, images, and descriptive captions—highlighting the benefit of incorporating multiple contextual sources.

To further enhance semantic understanding, researchers have explored the integration of external knowledge bases. KnowleNet~\citep{yue2023knowlenet}, for instance, leverages ConceptNet to inject prior knowledge and assesses cross-modal semantic similarity at both sample and word levels. Other recent models, such as RCLMuFN~\citep{wang2024rclmufn}, introduce relational context learning and multi-path fusion networks to improve generalization and robustness in sarcasm detection.

Despite these advances, most existing methods either rely on monolithic model architectures or simple cross-modal fusion, which often fail to capture the composite and context-dependent nature of sarcasm. 
In this context, our proposed Commander-GPT addresses these challenges by decomposing multimodal sarcasm detection into cognitively meaningful sub-tasks, each handled by a specialized agent under centralized coordination. This design aims to improve the performance of multimodal sarcasm detection.

 \begin{figure*}[t]
\centering
\includegraphics[width=5.6in]{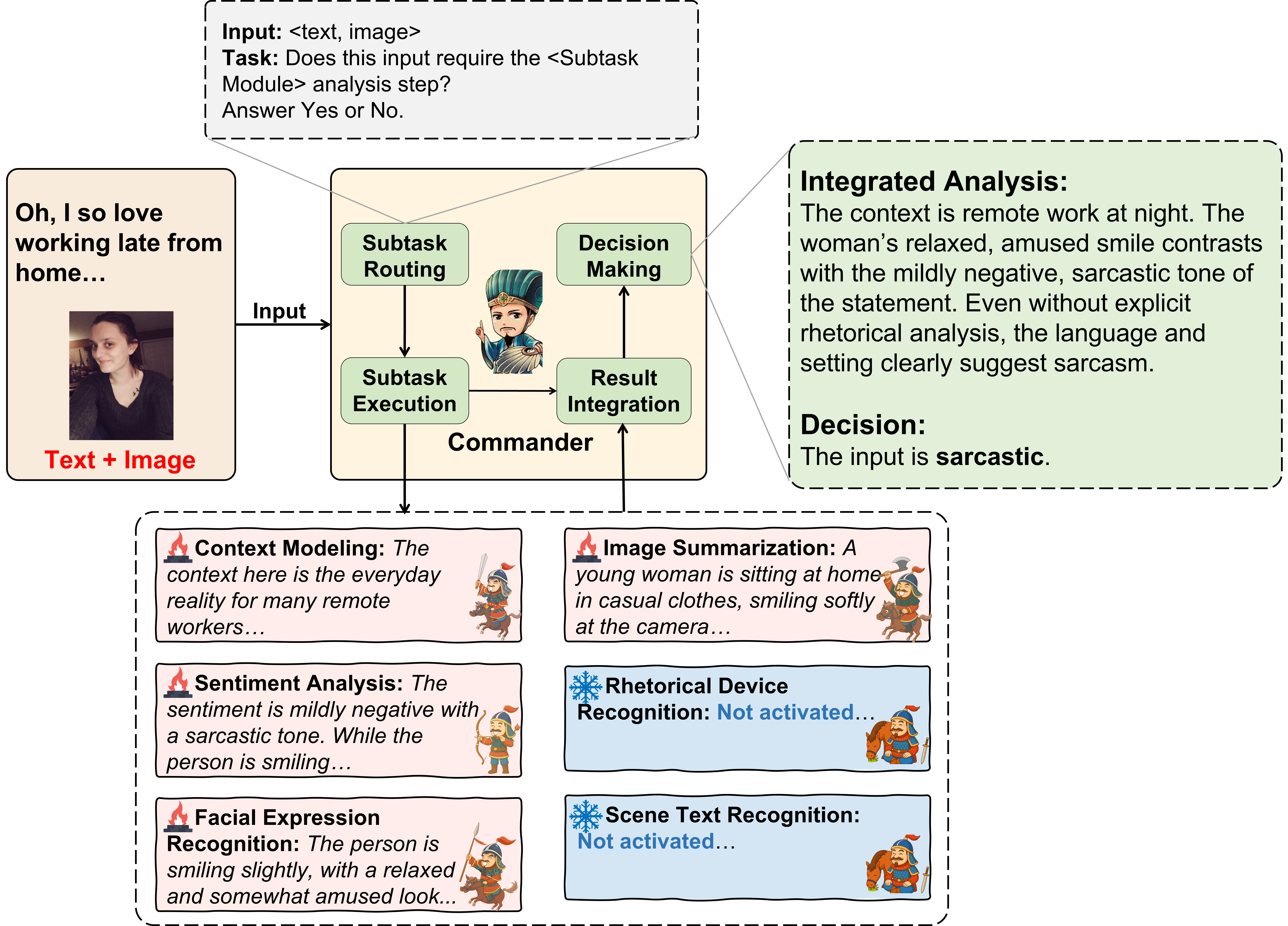}
\caption{The overall architecture of Commander-GPT.
}
\label{fig:model}
\end{figure*}

\section{The Proposed Approach}
To solve the limitation of current single MLLMs in understanding sarcasm, we propose Commander-GPT, a structured multi-agent framework that draws inspiration from military command structures where ``the right person for the right job'' principle ensures mission success. The overall architecture is shown in Fig.~\ref{fig:model}.

\subsection{Problem Formulation}
Consider a multimodal sarcasm detection task where we are given a dataset $\mathcal{D} = {(x_i, y_i)}_{i=1}^N$, where each instance $x_i = (t_i, v_i)$ consists of a textual component $t_i \in \mathcal{T}$ and a visual component $v_i \in \mathcal{V}$, and $y_i \in {0, 1}$ represents the binary sarcasm label. The objective is to learn a function $f: \mathcal{T} \times \mathcal{V} \rightarrow {0, 1}$ that accurately predicts sarcasm labels.

Traditional approaches employ a single monolithic model $\mathcal{M}$ to directly map the multimodal input to the prediction: $f(t, v) = \mathcal{M}(t, v)$. However, as established in prior work, this paradigm suffers from the inherent complexity of sarcasm understanding, which requires simultaneous processing of multiple linguistic and visual cues that may exhibit semantic contradictions.
In this work, we propose a fundamentally different approach inspired by military command structures. We decompose the complex sarcasm detection task into $K$ specialized subtasks, where each subtask $\tau_k$ for $k \in {1, 2, ..., K}$ focuses on extracting specific types of information. 

We assume access to a collection of expert models $\mathcal{A} = {A_1, A_2, ..., A_M}$, where each agent $A_j$ has varying capabilities across different subtasks. The capability of agent $A_j$ for subtask $\tau_k$ is characterized by a performance score $\pi_{j,k} \in [0, 1]$, where higher values indicate better suitability.
The core challenge lies in learning an optimal assignment function $\alpha: {1, 2, ..., K} \rightarrow {1, 2, ..., M}$ that maps each subtask $\tau_k$ to the most appropriate agent $A_{\alpha(k)}$. This assignment should maximize the overall system performance while considering the complementary strengths of different agents.

Given the assignment function $\alpha$, each subtask $\tau_k$ produces an intermediate output $z_k = \phi_k(x_i)$ through the assigned agent $A_{\alpha(k)}$. The final prediction is made by a commander model $\mathcal{C}$ that aggregates all intermediate representations:
\begin{equation}
f(t_i, v_i) = \mathcal{C}({z_k}_{k=1}^K) = \mathcal{C}(\phi_1(t_i), \phi_2(t_i), \phi_3(t_i), \phi_4(v_i), \phi_5(v_i), \phi_6(v_i))
\end{equation}

The commander model $\mathcal{C}$ serves as the central decision-making unit, analogous to a military command center that processes intelligence reports from specialized units and makes strategic decisions. This hierarchical architecture enables specialized processing while maintaining global coordination through the commander's integrative reasoning capabilities.
The optimization objective is to learn both the assignment function $\alpha$ and the commander model parameters $\theta_{\mathcal{C}}$ that minimize the expected sarcasm detection error:
\begin{equation}
\min_{\alpha, \theta_{\mathcal{C}}} \mathbb{E}{(x,y) \sim \mathcal{D}} \left[ \mathcal{L}(y, f(x; \alpha, \theta{\mathcal{C}})) \right]
\end{equation}
where $\mathcal{L}$ represents the binary cross-entropy loss function. This formulation captures the essence of our military-inspired approach: strategic task decomposition, specialized agent deployment, and centralized command coordination.

\subsection{Subtask Division}
Sarcasm understanding often involves: (1) understanding the broader context and pragmatic implications, (2) recognizing the surface-level emotional expression, (3) detecting linguistic markers such as irony and hyperbole, (4) interpreting visual emotional cues that may contradict textual sentiment, (5) understanding the visual scene context, and (6) processing any textual information embedded in images.
Based on this theoretical foundation, we decompose the complex sarcasm detection task into $K = 6$ cognitively meaningful subtasks, where each subtask $\tau_k$ focuses on extracting specific types of information that contribute to sarcasm understanding:
\begin{itemize}
\item $\tau_1$: \textbf{Context Modeling}: Analyze broader conversational context and pragmatic implications from text $t$: $\phi_1(t) \rightarrow \mathcal{P}$, where $\mathcal{P}$ denotes the pragmatic interpretation space.
\item $\tau_2$: \textbf{Sentiment Analysis}: Extract fine-grained emotional polarity from text $t$: $\phi_2(t) \rightarrow \mathcal{S}$, where $\mathcal{S} \in \mathbb{R}^{28}$ represents multi-dimensional emotion categories (e.g., joy, anger, surprise). This identifies the surface-level emotional expression.
\item $\tau_3$: \textbf{Rhetorical Device Recognition}: Identify linguistic patterns and rhetorical structures in text $t$: $\phi_3(t) \rightarrow \mathcal{R}$, where $\mathcal{R}$ represents rhetorical markers such as irony, hyperbole, and understatement.
\item $\tau_4$: \textbf{Facial Expression Recognition}: Extract facial emotions from image $v$: $\phi_4(v) \rightarrow \mathcal{E}$, where $\mathcal{E} \in \mathbb{R}^{7}$ represents basic emotion categories (happy, sad, angry, fear, surprise, disgust, neutral).
\item $\tau_5$: \textbf{Image Summarization}: Generate comprehensive scene description from image $v$: $\phi_5(v) \rightarrow \mathcal{D}$, where $\mathcal{D}$ represents natural language descriptions of visual content and scene context.
\item $\tau_6$: \textbf{Scene Text Recognition}: Extract textual content embedded in image $v$: $\phi_6(v) \rightarrow \mathcal{T}{scene}$, where $\mathcal{T}{scene}$ represents text sequences found in the visual scene.
\end{itemize}
This decomposition is motivated by the observation that sarcasm often emerges from contradictions between these different information layers. For instance, a text expressing positive sentiment ($\tau_2$) accompanied by a person's annoyed facial expression ($\tau_4$) in a chaotic scene ($\tau_5$) may indicate sarcastic intent. By explicitly modeling these components, our framework can capture the nuanced interplay that previous approaches often miss.

\subsection{Routing Approach}

The routing scorer is a core component in Commander-GPT, enabling the commander to dynamically select relevant sub-task agents for a given input $x = (t, v)$. We provide two representative instantiations: a learnable classifier (e.g., BERT-based commander) and a prompt-based approach for LLMs (e.g., GPT-4o, DeepSeek VL).

\paragraph{Learnable Routing Classifier (Multimodal BERT based Commander).}

Let $K$ denote the total number of sub-tasks $\{\tau_k\}_{k=1}^K$. For each input $x = (t, v)$, we construct $K$ paired samples $(x, T_k)$, where $T_k$ is the descriptor for sub-task $\tau_k$ (e.g., ``sentiment analysis''). The routing scorer outputs an activation probability for each sub-task:
\begin{equation}
    p_k(x) = P_{\theta}(r_k = 1 \mid x, T_k) = \sigma\left(\mathbf{w}_k^\top h_{\mathrm{CLS}}(x, T_k) + b_k\right)
\end{equation}
where $h_{\mathrm{CLS}}(x, T_k) = \mathrm{Fusion}(\mathrm{BERT}(t),\, \mathrm{ViT}(v),\, T_k)$ denotes the fused multimodal representation obtained by combining the BERT-encoded text $t$, the ViT-encoded image $v$, and the sub-task descriptor $T_k$. Here, $\mathrm{Fusion}(\cdot)$ can be implemented as feature concatenation. The parameters $\mathbf{w}_k, b_k$ are trainable for each sub-task, and $\sigma$ is the sigmoid function.

To train the routing classifier, we require a labeled routing dataset $\mathcal{D}_{\mathrm{route}} = \{(x_i, T_k, r_{ik})\}$.  
We distill agent activation decisions from a powerful instruction-tuned vision-language model, GPT-4o. Specifically, for each training sample $x_i = (t_i, v_i)$ and each sub-task descriptor $T_k$, we prompt GPT-4o with a question such as:

\begin{tcolorbox}[colback=red!5!white, colframe=red!50!black, boxrule=0.8pt]
\textbf{Input}: <t\_i, v\_i> \\
\textbf{Task}: Does this input require the "<T\_k>" analysis step? \\
Answer "Yes" or "No".
\end{tcolorbox}

The binary response is recorded as the activation label $r_{ik} \in \{0, 1\}$. In this work, we distill approximately 5{,}000 routing supervision instances, enabling scalable and reliable training of the routing classifier.

The routing scorer is trained by minimizing the binary cross-entropy loss:
\begin{equation}
    \mathcal{L}_{\mathrm{route}} = - \frac{1}{N K} \sum_{i=1}^N \sum_{k=1}^K \left[ r_{ik} \log p_k(x_i) + (1 - r_{ik}) \log (1 - p_k(x_i)) \right]
\end{equation}

During inference, agent $A_k$ is activated for $x$ if $p_k(x) > \alpha_k$, where $\alpha_k \in [0, 1]$ is a tunable threshold.

\paragraph{Prompt-based Routing (LLM Commander).}

For instruction-tuned LLMs (e.g., GPT-4o, DeepSeek VL), agent routing is formulated as a natural language inference task. For each $(x, T_k)$, we construct the following prompt:
\begin{tcolorbox}[colback=red!5!white, colframe=red!50!black, boxrule=0.8pt]
\textbf{System:} You are a military commander analyzing intelligence to deploy specialized units for sarcasm detection.

\textbf{Input:} Text: "\{<t\_i>\}", Image: \{<v\_i>\}

\textbf{Task:} Determine which of the following units should be deployed:
\begin{itemize}
\item "context\_modeling": Analyze broader conversational context
\item "sentiment\_analysis": Extract emotional polarity
\item "rhetorical\_device": Identify irony, hyperbole, etc.
\item "facial\_expression": Analyze facial emotions in image
\item "image\_summarization": Describe visual content
\item "scene\_text": Extract text from image
\end{itemize}
\textbf{Output:} \{"context\_modeling": 0/1, "sentiment\_analysis": 0/1, ...\}
\end{tcolorbox}

The LLM $\mathcal{M}_{\mathrm{LLM}}$ is queried with $\mathcal{I}(x, T_k)$ and returns a textual response, which is mapped to $r_k(x) \in \{0, 1\}$ (1 if ``Yes'', 0 if ``No''). This approach requires no explicit training and leverages the zero-shot or few-shot abilities of modern LLMs.

\paragraph{Discussion.}
The BERT-based routing classifier is data-efficient, scalable, and can be trained jointly with the commander model $\mathcal{C}$. The prompt-based router utilizes the extensive knowledge of large LLMs, trading efficiency for generalization and flexibility.
Formally, the set of routed agents for input $x$ is:
\begin{equation}
    \mathcal{A}_x = \{ A_k \mid r_k(x) = 1,\ k=1,\ldots,K \}
\end{equation}

\subsection{Subtask Execution}

Once the routing scorer identifies the relevant sub-task agents for a given input $x_i = (t_i, v_i)$, each activated agent $A_k \in \mathcal{A}_{x_i}$ independently processes its assigned sub-task $\tau_k$. These specialized agents operate in parallel, leveraging their unique capabilities to extract specific types of information from the textual and visual components of $x_i$.

\paragraph{Linguistic Specialists.}
For the textual component $t_i$ of input $x_i$, three linguistic specialist agents are employed:

\begin{itemize}
    \item \textbf{Context Modeling Agent (Llama 3-8B):} This agent is responsible for analyzing the broader conversational context and pragmatic implications embedded within the text. It is invoked with a specific prompt to guide its analysis:
    \[
        z_1 = \phi_1(t_i) = \mathrm{Llama~3}(\text{``Analyze the contextual implications: ''} + t_i)
    \]
    The output $z_1$ captures the deeper meaning and situational context of the text.

    \item \textbf{Sentiment Analysis Agent (RoBERTa):} This agent extracts fine-grained sentiment polarity from the text using a fine-tuned RoBERTa model for sentiment analysis:
    \[
        z_2 = \phi_2(t_i) = \mathrm{RoBERTa}_{\text{emotions}}(t_i) \in \mathbb{R}^{3}
    \]
    The output $z_2$ is a 3-dimensional vector representing \textit{positive}, \textit{neutral} and \textit{negative}, providing a nuanced understanding of the text's surface-level emotional expression.

    \item \textbf{Rhetorical Device Agent (GLM-2B):} This agent identifies linguistic patterns and rhetorical structures, such as irony, hyperbole, or understatement, that are indicative of sarcasm. It uses a GLM-2B model optimized for keyword or pattern extraction:
    \[
        z_3 = \phi_3(t_i) = \mathrm{GLM\text{-}2B}_{\text{rhetoric}}(t_i)
    \]
    The output $z_3$ highlights key linguistic markers that contribute to sarcastic intent.
\end{itemize}

\paragraph{Visual Specialists.}
For the visual component $v_i$ of input $x_i$, three visual specialist agents are employed:

\begin{itemize}
    \item \textbf{Facial Expression Agent (ViT-FER):} This agent recognizes facial emotions present in the image and provides associated confidence scores, using a Vision Transformer (ViT) model trained for facial expression recognition:
    \[
        z_4 = \phi_4(v_i) = \mathrm{ViT\text{-}FER}(v_i) \in \mathbb{R}^{7}
    \]
    The output $z_4$ is a 7-dimensional vector representing basic emotion categories (e.g., happy, sad, angry), offering crucial visual emotional cues that might contradict textual sentiment.

    \item \textbf{Image Summarization Agent (BLIP-2):} This agent generates comprehensive natural language descriptions of the visual content and scene context:
    \[
        z_5 = \phi_5(v_i) = \mathrm{BLIP\text{-}2}(v_i) \rightarrow \text{natural language description}
    \]
    The output $z_5$ provides a high-level understanding of the visual scene, which is vital for contextualizing the overall message.

    \item \textbf{Scene Text Agent (OCR-2.0):} This agent extracts any textual content embedded within the image:
    \[
        z_6 = \phi_6(v_i) = \mathrm{OCR\text{-}2.0}(v_i) \rightarrow \text{extracted text sequence}
    \]
    The output $z_6$ provides additional textual information that might be a critical component of the sarcastic expression.
\end{itemize}

Each activated agent produces a structured output $z_k$ that typically includes both the extracted information and relevant confidence scores. The parallel execution of these agents ensures computational efficiency, while their specialized focus maintains the precise analysis required for complex sarcasm understanding.

\subsection{Result Integration}

After all activated agents have processed their respective sub-tasks, the Commander module $\mathcal{C}$ is responsible for integrating the intermediate outputs $\{z_k\}_{k \in \mathcal{A}_{\text{active}}}$ and making the final sarcasm prediction. 

\paragraph{Lightweight Encoder-Based Commander.}
For encoder-based commanders (e.g., multi-modal BERT), we concatenate the outputs of all activated agents and project them through a learned fusion and classification head:
\begin{align}
\mathbf{h}_{\text{fused}} &= \text{concat}([z_k\ \text{for}\ k \in \mathcal{A}_{\text{active}}]) \\
\mathbf{h}_{\text{context}} &= \mathrm{BERT}([\mathrm{CLS}; \mathbf{h}_{\text{fused}}; \mathrm{SEP}]) \\
\hat{y} &= \mathrm{softmax}(\mathbf{W}_{\text{out}} \mathbf{h}_{\text{context}} + \mathbf{b}_{\text{out}})
\end{align}
where $\mathbf{W}_{\text{out}} \in \mathbb{R}^{2 \times d_{\text{hidden}}}$ and $\mathbf{b}_{\text{out}} \in \mathbb{R}^2$ are trainable parameters for binary classification.

\paragraph{Autoregressive Language Model Commander.}
For moderately-sized language model commanders (e.g., DeepSeek-VL-7B), the outputs of all activated agents are formatted into a structured prompt using a predefined template:
\begin{align}
\mathbf{p}_{\text{input}} &= \text{Template}(\{z_k\}_{k \in \mathcal{A}_{\text{active}}}) \\
\hat{y} &= \mathcal{C}_{\text{LM}}(\mathbf{p}_{\text{input}}) \rightarrow \{\text{Sarcastic}, \text{Non-sarcastic}\}
\end{align}

\paragraph{Large LLM Commander.}
For large-scale LLM-based commanders (e.g., GPT-4o and Gemini Pro), all agent outputs are presented as a comprehensive structured prompt, and the LLM predicts both the sarcasm label and explanation:
\begin{tcolorbox}[colback=red!5!white, colframe=red!50!black, boxrule=0.8pt]
\textbf{System:} You are the supreme commander making the final sarcasm detection decision based on the reports.

\textbf{Reports:}

Context Analysis: \{$z_1$ if activated\}\\
Sentiment Analysis: \{$z_2$ if activated\}\\
Rhetorical Devices: \{$z_3$ if activated\}\\
Facial Expression: \{$z_4$ if activated\}\\
Image Description: \{$z_5$ if activated\}\\
Scene Text: \{$z_6$ if activated\}

\textbf{Task:} Analyze these reports and determine if the content is sarcastic. Consider contradictions between textual sentiment and visual cues, rhetorical patterns, and contextual incongruity.

\textbf{Output:} \{"prediction": "sarcastic"/"non-sarcastic"\}

\textbf{Reasoning:} \{Explanation\}
\end{tcolorbox}


This three-stage architecture ensures that our Commander-GPT framework maintains both the specialized expertise of individual agents and the strategic coordination capabilities of military command structures, leading to more accurate and interpretable sarcasm detection results.

\begin{algorithm}[ht]
\caption{Commander-GPT: Modular Multimodal Sarcasm Understanding}
\label{alg:commander_gpt}
\begin{algorithmic}[1]
\State \textbf{Input:} Multimodal input $x = (t, v)$
\State \textbf{Output:} Sarcasm prediction $\hat{y}$

\Statex
\State \textbf{1. Agent Routing:}
\State Obtain agent activation vector $[r_1(x), \dots, r_K(x)]$ using routing scorer.
\State $\mathcal{A}_x \gets \{A_k~|~r_k(x) = 1\}$

\Statex
\State \textbf{2. Subtask Execution:}
\ForAll{activated agent $A_k \in \mathcal{A}_x$ \textbf{ in parallel}}
    \State Compute subtask output $z_k = \phi_k(x)$
\EndFor
\State Collect all outputs $\mathbf{z} = [z_k~|~A_k \in \mathcal{A}_x]$

\Statex
\State \textbf{3. Result Integration and Final Decision:}
\If{commander is encoder-based}
    \State Fuse outputs: $\mathbf{h}_{\text{fused}} \gets \mathrm{concat}(\mathbf{z})$
    \State Compute hidden: $\mathbf{h}_{\text{context}} \gets \mathrm{BERT}([\mathrm{CLS}; \mathbf{h}_{\text{fused}}; \mathrm{SEP}])$
    \State Predict: $\hat{y} = \mathrm{softmax}(\mathbf{W}_{\text{out}} \mathbf{h}_{\text{context}} + \mathbf{b}_{\text{out}})$
\Else
    \State Format prompt $\mathbf{p}_{\text{input}}$ from all $z_k$
    \State Predict: $\hat{y} = \mathcal{C}_{\text{LLM}}(\mathbf{p}_{\text{input}})$
\EndIf

\end{algorithmic}
\end{algorithm}




\section{Experiments}

In this section, we conduct comprehensive experiments on two widely-used multimodal sarcasm detection benchmarks, MMSD~\citep{cai2019multi} and MMSD 2.0~\citep{qin2023mmsd2}. Our results show that Commander-GPT consistently outperforms state-of-the-art baselines, thereby demonstrating the effectiveness and generalizability of our proposed framework.

\subsection{Datasets}

\textbf{MMSD:} The MMSD dataset is a benchmark for multimodal sarcasm detection, comprising paired textual and visual data collected from Twitter. Each example consists of a tweet and its associated image, with ground-truth sarcasm annotations. The textual content frequently features subtle or implicit sarcasm, while the accompanying images provide additional context, making the task especially challenging for both unimodal and multimodal models.

\textbf{MMSD 2.0:} MMSD 2.0 is an enhanced and extended version of MMSD, designed to support more robust evaluation of multimodal sarcasm detection systems. Compared to its predecessor, MMSD 2.0 significantly increases both the diversity of visual content and the quality of text-image alignment. It introduces more challenging cases where understanding sarcasm requires reasoning over complex interactions between modalities. Key statistics and properties of the two datasets are summarized in Table~\ref{tab:dataset}.

\begin{table*}[t]
\centering
\small
\setlength{\tabcolsep}{8pt}
\renewcommand{\arraystretch}{1.12}
\caption{Statistics of the MMSD and MMSD 2.0 datasets.}
\label{tab:dataset}
\begin{tabular}{lcccccc}
\toprule
\textbf{Dataset} & \textbf{Train} & \textbf{Validation} & \textbf{Test} & \textbf{Sarcastic} & \textbf{Non-sarcastic} & \textbf{Source} \\
\midrule
MMSD     & 19,816 & 2,410 & 2,409 & 10,560 & 14,075 & Twitter \\
MMSD 2.0 & 19,816 & 2,410 & 2,409 & 11,651 & 12,980 & Twitter \\
\bottomrule
\end{tabular}
\end{table*}

\subsection{Experimental Settings}

\paragraph{Implementation Details.}
All experiments are conducted on a server equipped with two NVIDIA RTX 4090 GPUs and 256GB RAM. The Commander-GPT framework is implemented using PyTorch, HuggingFace Transformers, and OpenMMLab toolkits. For models requiring supervised training (e.g., BERT-based commander and routing classifier), we use a batch size of 64, a maximum sequence length of 512, and the Adam optimizer with an initial learning rate of $2\times10^{-5}$. All hyperparameters are tuned on the validation set unless otherwise specified. For LLMs, we only perform inference without any parameter tuning or fine-tuning.

\paragraph{Evaluation Metrics.}
We use four standard metrics for performance evaluation: accuracy (Acc.), precision (P), recall (R), and F1-score. Among these, F1-score serves as the primary criterion, as it provides a balanced assessment of precision and recall, which is especially important for imbalanced class distributions.

\paragraph{Baselines and Commander Configurations.}
We compare Commander-GPT against five representative prompting and reasoning baselines:
\begin{itemize}
    \item \textbf{Plan-and-Solve}~\citep{wang2023plan}: A pipeline prompting strategy that first plans the solution steps and then solves each subproblem sequentially.
    \item \textbf{Zero-shot CoT}~\citep{kojima2022large}: A zero-shot chain-of-thought prompting method that enables step-by-step reasoning without requiring labeled demonstrations.
    \item \textbf{Generated Knowledge Prompting}~\citep{liu2021generated}: A method that augments the input with external knowledge generated by a language model to enhance reasoning.
    \item \textbf{Automatic Prompt Engineer}~\citep{zhou2022large}: A technique that automatically searches for and optimizes prompt templates to maximize downstream task performance.
    \item \textbf{S$^3$ Agent}~\citep{wang2024s3}: A multi-agent coordination framework for complex task decomposition and solution synthesis.
\end{itemize}
We further evaluate Commander-GPT under seven commander configurations, ranging from lightweight encoder-based commanders (BERT), to four medium-sized autoregressive LLMs (Yi-VL (6B)~\citep{young2024yi}, DeepSeek-VL-Chat (7B)~\citep{lu2024deepseek}, Qwen-VL-Chat (9B)~\citep{bai2023qwen}, and MiniCPM-V-2 (2.8B)~\citep{hu2024minicpm}), and two large instruction-following LLMs (Gemini Pro and GPT-4o).

\subsection{Main results}
Table~\ref{tab:main_results} summarizes the performance of different commander architectures at three scales: lightweight encoder-based, small open-source LLMs, and SoTA proprietary LLMs. For all experiments, our routing-based method achieves the highest F1 scores and provides significant improvements over corresponding baselines.

\textbf{Lightweight Encoder-Based Commander.}  
The BERT+ViT-based commander represents the lightweight solution with minimal model size (110M parameters). Compared to direct fine-tuning, our routing method achieves substantial absolute gains: on MMSD, F1 improves from 80.8 to 86.7 (\textbf{+5.9}), and on MMSD 2.0 from 77.2 to 85.8 (\textbf{+8.6}). This corresponds to relative improvements of 7.3\% and 11.1\%, respectively. The results indicate that even for parameter-efficient models with limited multimodal reasoning ability, our approach enables more effective use of learned features, significantly boosting both overall accuracy and robustness.

\textbf{Small Autoregressive LLMs Commander.}  
For the small open-source LLMs group (Yi-VL, DeepSeek-VL-Chat, Qwen-VL-Chat, MiniCPM-V2), our method consistently outperforms all competitive prompting and multi-agent baselines. On MMSD 2.0, F1 improvements are especially pronounced: \textbf{+23.6} (Yi-VL), \textbf{+8.5} (DeepSeek-VL-Chat), \textbf{+10.3} (Qwen-VL-Chat), and \textbf{+22.8} (MiniCPM-V2). For instance, Qwen-VL-Chat reaches 68.9 F1 versus 58.6 for the best prior method, and MiniCPM-V2 rises from 49.2 to 72.0 F1. These results suggest that our approach is highly effective at extracting complementary information from multiple sub-agents in smaller models, compensating for their individual limitations.

\textbf{SoTA LLMs Commander.}  
For the strongest closed-source models (Gemini Pro, GPT-4o), the routing approach continues to yield improvements, though the margins are reduced compared to smaller models due to the high baseline performance. On MMSD 2.0, Gemini Pro improves from 67.4 to 71.9 F1 (\textbf{+4.5}), and GPT-4o increases from 73.2 to 76.5 F1 (\textbf{+3.3}). This demonstrates that even state-of-the-art models, which already possess advanced multi-modal capabilities, benefit from structured subtask decomposition and information routing.

\textbf{Summary.}  
In conclusion, our method consistently improves F1 performance across all commander types, with the most dramatic relative gains observed in lightweight and small open-source models. The gap in improvement narrows as the model scale increases, suggesting diminishing returns for extremely strong baselines. Nevertheless, the overall trend validates the generality and scalability of our routing architecture, establishing its effectiveness across a wide range of model sizes and types.


\begin{table*}[t]
\caption{Main Experimental Results. The \textbf{bolded} numbers indicate the best performance. Additionally, we calculated the increase in the F1 score.}
\label{tab:main_results}
\centering
\scalebox{0.75}{
\begin{tabular}{cclcccccccc}
\toprule
                                   &                                                   &                               & \multicolumn{4}{c}{\textbf{MMSD}}                                                                                                                                                                   & \multicolumn{4}{c}{\textbf{MMSD 2.0}}                                                                                                                                         \\ \cline{4-11} 
\multirow{-2}{*}{\textbf{Model}}   & \multirow{-2}{*}{\textbf{Parameters}} & \multirow{-2}{*}{\textbf{Method}}      & \textbf{F1}                                                               & \textbf{Acc.}                                  & \textbf{Pre.}                                  & \textbf{Rec.}                                  & \textbf{F1}                                          & \textbf{Acc.}                                  & \textbf{Pre.}                                  & \textbf{Rec.}                                  \\ \hline
\multicolumn{11}{c}{\cellcolor[HTML]{F0F0F0}\textbf{\large Lightweight Encoder-Based Commander}} \\ \hline
                                                               &                        & BERT (Fine-tuned)             & 80.8                                                               & 82.6                                  & 81.2                                  & 80.1                                  & 77.2                                         & 79.7                                  & 77.4                                  & 75.8                                  \\
\multirow{-2}{*}{BERT+ViT}        & \multirow{-2}{*}{110M}       & \cellcolor[HTML]{E8F4F8}{\textbf{Ours}}  & \cellcolor[HTML]{E8F4F8}{ \textbf{86.7 (5.9↑)}}                        & \cellcolor[HTML]{E8F4F8}{ \textbf{90.1}} & \cellcolor[HTML]{E8F4F8}{ \textbf{87.4}} & \cellcolor[HTML]{E8F4F8}{ \textbf{86.1}} & \cellcolor[HTML]{E8F4F8}{ \textbf{85.8 (8.6↑)}}  & \cellcolor[HTML]{E8F4F8}{ \textbf{87.4}} & \cellcolor[HTML]{E8F4F8}{ \textbf{86.1}} & \cellcolor[HTML]{E8F4F8}{ \textbf{85.3}} \\ \toprule

\multicolumn{11}{c}{\cellcolor[HTML]{F0F0F0}\textbf{\large Small Autoregressive LLMs Commander}} \\ \hline
                                   &                                                    & Zero-shot CoT                 & 45.8                                                               & 59.2                                  & 51.4                                  & 41.4                                  & 6.1                                          & 16.2                                  & 6.0                                   & 6.3                                   \\
                                   
                                   &                                                    & Automatic Prompt Engineer     & 49.4                                                               & 53.3                                  & 45.0                                  & 54.7                                  & 4.3                                          & 14.2                                  & 4.2                                   & 4.5                                   \\

                                   &                                                     & Plan-and-Solve                & 51.3                                                               & 62.6                                  & 56.2                                  & 47.2                                  & 4.4                                          & 9.2                                   & 4.0                                   & 4.8                                   \\
                                   
                                   &                                                     & Generated Knowledge Prompting & 52.7                                                               & 55.7                                  & 47.5                                  & 59.2                                  & 4.3                                          & 11.2                                  & 4.0                                   & 4.6                                   \\
                                   
                                   &                                                   & $S^3$ Agent                   & 52.3                                                               & 41.4                                  & 39.6                                  & \textbf{77.0}                         & 31.7                                         & 54.3                                  & 44.4                                  & 24.6                                  \\

\multirow{-6}{*}{Yi-VL}            & \multirow{-6}{*}{6B}        & \cellcolor[HTML]{E8F4F8}{ \textbf{Ours}}  & \cellcolor[HTML]{E8F4F8}{ \textbf{59.1 (6.4↑)}}                        & \cellcolor[HTML]{E8F4F8}{ \textbf{70.8}} & \cellcolor[HTML]{E8F4F8}{ \textbf{71.2}} & \cellcolor[HTML]{E8F4F8}{ 50.5}          & \cellcolor[HTML]{E8F4F8}{ \textbf{55.3 (23.6↑)}} & \cellcolor[HTML]{E8F4F8}{ \textbf{67.9}} & \cellcolor[HTML]{E8F4F8}{ \textbf{69.0}} & \cellcolor[HTML]{E8F4F8}{ \textbf{46.1}} \\ \cline{3-11} 

                                   &                                                 & Zero-shot CoT                 & 54.4                                                               & 65.3                                  & 60.1                                  & 49.8                                  & 48.8                                         & 62.8                                  & 59.8                                  & 41.3                                  \\
                                   
                                   &                                                   & Automatic Prompt Engineer     & 55.1                                                               & 68.0                                  & \textbf{66.4}                         & 47.6                                  & 46.6                                         & 62.3                                  & 59.7                                  & 38.2                                  \\
                                   
                                   &                                                   & Plan-and-Solve                & 54.9                                                               & 61.9                                  & 54.3                                  & 55.5                                  & 48.3                                         & 63.0                                  & 60.5                                  & 40.2                                  \\

                                   &                                                    & Generated Knowledge Prompting & 55.6                                                               & 56.8                                  & 48.7                                  & \textbf{65.0}                         & 27.5                                         & 58.1                                  & 53.8                                  & 18.5                                  \\
                                   
                                   &                                                    & $S^3$ Agent                   & 59.7                                                               & 45.3                                  & 43.1                                  & 59.7                                  & 52.0                                         & \textbf{64.9}                         & \textbf{63.3}                         & 44.1                                  \\
\multirow{-6}{*}{DeepSeek-VL-Chat} & \multirow{-6}{*}{7B}         & \cellcolor[HTML]{E8F4F8}{ \textbf{Ours}}  & \cellcolor[HTML]{E8F4F8}{ \textbf{61.1 (1.4↑)}}                        & \cellcolor[HTML]{E8F4F8}{ \textbf{69.4}} & \cellcolor[HTML]{E8F4F8}{ 64.9}          & \cellcolor[HTML]{E8F4F8}{ 57.7}          & \cellcolor[HTML]{E8F4F8}{ \textbf{60.5 (8.5↑)}}  & \cellcolor[HTML]{E8F4F8}{ 46.7}          & \cellcolor[HTML]{E8F4F8}{ 44.4}          & \cellcolor[HTML]{E8F4F8}{ \textbf{94.7}} \\ \cline{3-11}

                                   &                                                 & Zero-shot CoT                 & 66.4                                                               & 66.0                                  & 56.5                                  & 80.7                                  & 33.6                                         & 40.0                                  & 32.0                                  & 35.3                                  \\
                                   
                                   &                                                    & Automatic Prompt Engineer     & 64.5                                                               & 60.5                                  & 51.6                                  & 86.0                                  & 33.3                                         & 40.4                                  & 32.1                                  & 34.5                                  \\
                                   
                                   &                                                    & Plan-and-Solve                & 59.4                                                               & 64.8                                  & 57.2                                  & 61.8                                  & 34.8                                         & 38.7                                  & 32.1                                  & 38.0                                  \\
                                   
                                   &                                                    & Generated Knowledge Prompting & 50.9                                                               & 67.1                                  & \textbf{64.7}                         & 40.9                                  & 28.0                                         & 38.9                                  & 38.4                                  & 37.6                                  \\
                                   
                                   &                                                    & $S^3$ Agent                   & 68.1                                                               & 67.5                                  & 57.6                                  & 83.3                                  & 58.6                                         & 63.3                                  & 57.0                                  & 60.4                                  \\
                                   
\multirow{-6}{*}{Qwen-VL-Chat}     & \multirow{-6}{*}{9B}        & \cellcolor[HTML]{E8F4F8}{ \textbf{Ours}}  & \cellcolor[HTML]{E8F4F8}{ \textbf{69.3 (1.2↑)}}                        & \cellcolor[HTML]{E8F4F8}{ \textbf{68.2}} & \cellcolor[HTML]{E8F4F8}{ 58.0}          & \cellcolor[HTML]{E8F4F8}{ \textbf{86.2}} & \cellcolor[HTML]{E8F4F8}{ \textbf{68.9 (10.3↑)}} & \cellcolor[HTML]{E8F4F8}{ \textbf{67.7}} & \cellcolor[HTML]{E8F4F8}{ \textbf{58.8}} & \cellcolor[HTML]{E8F4F8}{ \textbf{83.2}} \\ \cline{3-11}

                                   &                                                   & Zero-shot CoT                 & 61.4                                                               & 68.1                                  & 62.1                                  & 60.7                                  & 33.0                                         & 51.6                                  & 40.9                                  & 27.7                                  \\
                                   
                                   &                                                     & Automatic Prompt Engineer     & 63.2                                                               & 69.2                                  & 63.1                                  & 63.4                                  & 31.2                                         & 48.0                                  & 36.2                                  & 27.4                                  \\
                                   
                                   &                                                   & Plan-and-Solve                & 60.9                                                               & 66.0                                  & 58.5                                  & 63.5                                  & 37.7                                         & 46.8                                  & 38.0                                  & 37.4                                  \\
                                   
                                   &                                            & Generated Knowledge Prompting & 63.2                                                               & 67.5                                  & 59.9                                  & 66.8                                  & 47.5                                         & 47.3                                  & 41.6                                  & 55.3                                  \\
                                   
                                   &                                                   & $S^3$ Agent                   & 60.8                                                               & 66.6                                  & 59.6                                  & 62.0                                  & 49.2                                         & 62.9                                  & 60.0                                  & 41.8                                  \\
                                   
\multirow{-6}{*}{MiniCPM-V2}      & \multirow{-6}{*}{2.8B}       & \cellcolor[HTML]{E8F4F8}{ \textbf{Ours}}  & \cellcolor[HTML]{E8F4F8}{ \textbf{72.5 (9.3↑)}} & \cellcolor[HTML]{E8F4F8}{ \textbf{74.5}} & \cellcolor[HTML]{E8F4F8}{ \textbf{65.9}} & \cellcolor[HTML]{E8F4F8}{ \textbf{80.8}} & \cellcolor[HTML]{E8F4F8}{ \textbf{72.0 (22.8↑)}} & \cellcolor[HTML]{E8F4F8}{ \textbf{73.9}} & \cellcolor[HTML]{E8F4F8}{ \textbf{66.8}} & \cellcolor[HTML]{E8F4F8}{ \textbf{78.1}} \\ \toprule

\multicolumn{11}{c}{\cellcolor[HTML]{F0F0F0}\textbf{\large SoTA LLMs Commander}} \\ \hline
                                   &                                                  & Zero-shot CoT                 & 67.2                                                               & 71.4                                 & 62.8                                 & 70.5                                  & 59.4                                         & 61.1                                  & 53.9                                  &66.1                                 \\
                                   
                                   &                                                  & Automatic Prompt Engineer     & 68.7                                                               & 72.8                                 & 64.1                                 & 73.9                                  & 60.0                                         &60.4                                  & 53.1                                  &68.1                                    \\
                                   
                                   &                                                   & Plan-and-Solve               & 66.3                                                               & 70.2                                 & 62.5                                 & 71.8                                  & 53.7                                         & 56.5                                  & 49.6                                  &58.5                                   \\
                                   
                                   &                                                   & Generated Knowledge Prompting & 69.5                                                               & 73.6                                 & 66.3                                 & 74.2                                  & 65.1                                         & 61.7                                  & 53.6                                  &\textbf{82.8}                                   \\
                                   
                                   &                                                   & $S^3$ Agent                  & 70.1                                                               & 74.9                                 & 67.8                                 & 75.7                                  & 67.4                                         & 66.5                                  & 58.0                                  &80.3                                    \\
                                   
\multirow{-6}{*}{Gemini Pro}           & \multirow{-6}{*}{600B}          & \cellcolor[HTML]{E8F4F8}{ \textbf{Ours}}  & \cellcolor[HTML]{E8F4F8}{ \textbf{73.8 (3.7↑)}}                        & \cellcolor[HTML]{E8F4F8}{ \textbf{76.2}} & \cellcolor[HTML]{E8F4F8}{ \textbf{70.1}} & \cellcolor[HTML]{E8F4F8}{ \textbf{78.3}} & \cellcolor[HTML]{E8F4F8}{ \textbf{71.9 (4.5↑)}}  & \cellcolor[HTML]{E8F4F8}{ \textbf{73.8}} & \cellcolor[HTML]{E8F4F8}{ \textbf{68.4}} & \cellcolor[HTML]{E8F4F8}{ 76.7} \\\cline{3-11}

                                   &                                                  & Zero-shot CoT                 & 74.2                                                               & 78.5                                  & 71.3                                  & 77.4                                  & 68.9                                         & 74.2                                  & 65.8                                  & 72.6                                  \\
                                   
                                   &                                                  & Automatic Prompt Engineer     & 75.8                                                               & 79.1                                  & 72.9                                  & 78.9                                  & 70.3                                         & 75.6                                  & 67.1                                  & 73.8                                  \\
                                   
                                   &                                                   & Plan-and-Solve                & 73.6                                                               & 77.8                                  & 70.2                                  & 77.1                                  & 67.4                                         & 73.1                                  & 64.2                                  & 71.0                                  \\
                                   
                                   &                                                   & Generated Knowledge Prompting & 76.3                                                               & 79.7                                  & 73.5                                  & 79.2                                  & 71.8                                         & 76.4                                  & 68.9                                  & 75.1                                  \\
                                   
                                   &                                                   & $S^3$ Agent                   & 77.9                                                               & 80.2                                  & 74.6                                  & 81.5                                  & 73.2                                         & 77.8                                  & 70.3                                  & 76.4                                  \\
                                   
\multirow{-6}{*}{GPT-4o}           & \multirow{-6}{*}{300B}          & \cellcolor[HTML]{E8F4F8}{ \textbf{Ours}}  & \cellcolor[HTML]{E8F4F8}{ \textbf{81.4 (3.5↑)}}                        & \cellcolor[HTML]{E8F4F8}{ \textbf{83.1}} & \cellcolor[HTML]{E8F4F8}{ \textbf{78.9}} & \cellcolor[HTML]{E8F4F8}{ \textbf{84.2}} & \cellcolor[HTML]{E8F4F8}{ \textbf{76.5 (3.3↑)}}  & \cellcolor[HTML]{E8F4F8}{ \textbf{79.7}} & \cellcolor[HTML]{E8F4F8}{ \textbf{74.6}} & \cellcolor[HTML]{E8F4F8}{ \textbf{79.4}} \\ \toprule
\end{tabular}
}
\end{table*}

 \begin{figure}[t]
    \centering
    \includegraphics[width=3.5in]{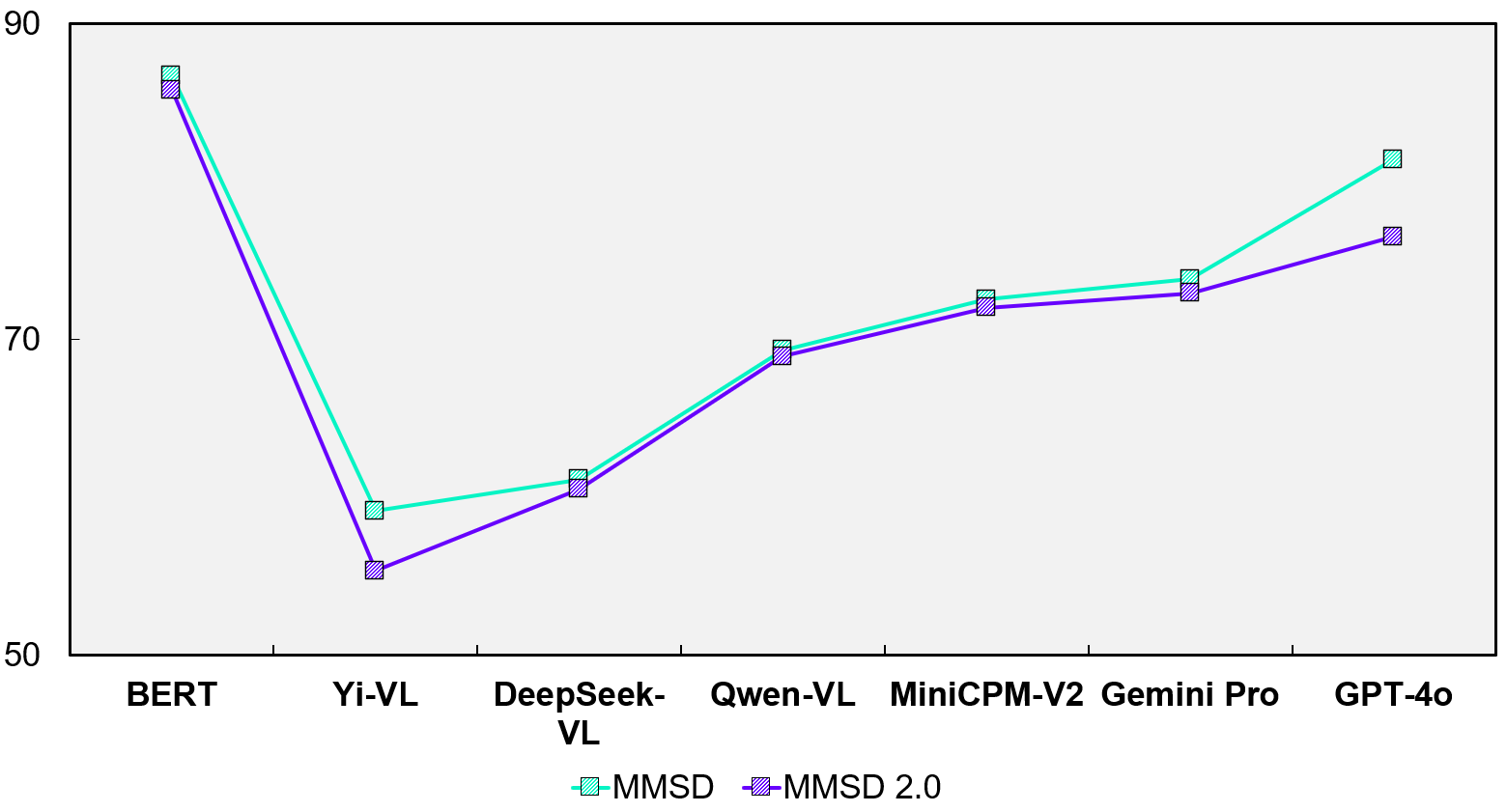}
    \caption{Comparison of commander models on the MMSD and MMSD 2.0 dataset.}
   \label{fig:commander_comparison}
\end{figure}

\subsection{Comparison of Commander Models}
Fig.~\ref{fig:commander_comparison} presents the F1 scores of various commander models on the MMSD and MMSD 2.0 datasets. The results reveal significant performance differences among LLM backbones. BERT achieves the highest F1 scores (86.7 on MMSD, 85.8 on MMSD 2.0), primarily due to its fine-tuned nature on the sarcasm detection task. In contrast, mainstream vision-language models, such as Yi-VL (59.1/55.3) and DeepSeek-VL (61.1/60.5), perform substantially worse, likely due to their weaker language modeling capability or suboptimal adaptation to sarcasm-rich domains. Larger multimodal models like Qwen-VL, MiniCPM-V2, Gemini Pro, and GPT-4o show a clear performance hierarchy: GPT-4o delivers the strongest results among MLLMs (81.4/76.5), benefiting from both superior language modeling and robust multi-modal reasoning. Overall, these results highlight the importance of both model architecture and scale for effective sarcasm detection, and demonstrate that Commander-GPT can flexibly leverage a wide range of LLMs, with significant performance gain observed as model quality improves.

\subsection{Ablation Study}

Table~\ref{tab:ablation-mmsd} and Table~\ref{tab:ablation-mmsd2} present the ablation results on the MMSD and MMSD 2.0 datasets, respectively. Across all models, removing any single sub-task module leads to a noticeable drop in both F1 and accuracy (\%), confirming the necessity of each component. For all three models, the largest relative performance degradation occurs when either \textit{Rhetorical Device Recognition} or \textit{Context Modeling} is ablated. For example, on MMSD, removing Rhetorical Device Recognition results in a 24.7\% F1 decrease for DeepSeek-VL (from 61.1\% to 27.9\%), a 7.2\% drop for MiniCPM-V2, and a 5.2\% decrease for GPT-4o. Removing Context Modeling leads to a 6.9\% F1 drop for DeepSeek-VL, 0.1\% for MiniCPM-V2, and 2.5\% for GPT-4o. On MMSD 2.0, similar trends are observed, with the ablation of Rhetorical Device Recognition causing up to 29.1\% relative F1 drop for DeepSeek-VL, and 3.5\% for GPT-4o.

For other sub-tasks such as Sentiment Analysis, Facial Expression Recognition, Image Summarization, and Scene Text Recognition, the impact is present but less pronounced, with typical F1 reductions ranging from 0.2\% to 1.1\% for GPT-4o, and 0.5\% to 1.3\% for MiniCPM-V2. All models achieve their best F1 and accuracy when all six modules are integrated, validating the effectiveness and complementarity of the full routing strategy. Overall, these results demonstrate that multi-dimensional sub-task collaboration is critical for robust multi-modal sarcasm detection, with context and rhetorical information providing the most substantial performance gains.

\begin{table*}[t]
\caption{Ablation study on the MMSD dataset.}
\label{tab:ablation-mmsd}
\centering
\small
\begin{tabular}{lcc|cc|cc}
\toprule
\multirow{2}{*}{\textbf{Ablation Setting}} 
    & \multicolumn{2}{c|}{\textbf{MiniCPM-V2}} 
    & \multicolumn{2}{c|}{\textbf{DeepSeek-VL}} 
    & \multicolumn{2}{c}{\textbf{GPT-4o}} \\\cline{2-7}
    & F1 & Acc. & F1 & Acc. & F1 & Acc. \\
\midrule
w/o Context Modeling              & 72.4 & 73.9 & 56.9 & 70.9 & 78.9 & 81.2 \\
w/o Sentiment Analysis            & 71.3 & 72.2 & 60.2 & 71.9 & 79.1 & 81.4 \\
w/o Rhetorical Device Recognition & 66.0 & 72.4 & 27.9 & 63.9 & 77.2 & 79.8 \\
w/o Facial Expression Recognition & 71.2 & 72.2 & 41.7 & 66.8 & 78.5 & 81.0 \\
w/o Image Summarization           & 71.4 & 73.1 & 43.8 & 67.3 & 78.8 & 81.3 \\
w/o Scene Text Recognition        & 69.4 & 71.5 & 28.8 & 63.4 & 77.8 & 80.1 \\
\rowcolor[HTML]{E8F4F8}
{\textbf{Full Model}} & {\textbf{72.5}} & {\textbf{74.5}} & {\textbf{61.1}} & {\textbf{69.4}} & {\textbf{81.4}} & {\textbf{83.1}} \\
\bottomrule
\end{tabular}
\end{table*}

\begin{table*}[t]
\caption{Ablation study on the MMSD 2.0 dataset.}
\label{tab:ablation-mmsd2}
\centering
\small
\begin{tabular}{lcc|cc|cc}
\toprule
\multirow{2}{*}{\textbf{Ablation Setting}} 
    & \multicolumn{2}{c|}{\textbf{MiniCPM-V2}} 
    & \multicolumn{2}{c|}{\textbf{DeepSeek-VL}} 
    & \multicolumn{2}{c}{\textbf{GPT-4o}} \\\cline{2-7}
    & F1 & Acc. & F1 & Acc. & F1 & Acc. \\
\midrule
w/o Context Modeling              & 72.1 & 73.6 & 59.4 & 43.1 & 74.1 & 77.2 \\
w/o Sentiment Analysis            & 71.2 & 71.6 & 59.0 & 42.9 & 74.3 & 77.4 \\
w/o Rhetorical Device Recognition & 67.9 & 72.3 & 60.0 & 42.5 & 73.8 & 77.0 \\
w/o Facial Expression Recognition & 71.5 & 72.4 & 59.7 & 43.2 & 74.0 & 77.1 \\
w/o Image Summarization           & 71.9 & 73.3 & 59.5 & 43.0 & 74.2 & 77.3 \\
w/o Scene Text Recognition        & 67.9 & 62.2 & 60.2 & 43.1 & 73.2 & 76.4 \\
\rowcolor[HTML]{E8F4F8}
{\textbf{Full Model}} & {\textbf{72.0}} & {\textbf{73.9}} & {\textbf{60.5}} & {\textbf{46.7}} & {\textbf{76.5}} & {\textbf{79.7}} \\
\bottomrule
\end{tabular}
\end{table*}

\subsection{Subtask Invocation Frequency Analysis}
\begin{figure}[t]
    \centering
    \includegraphics[width=2.8in]{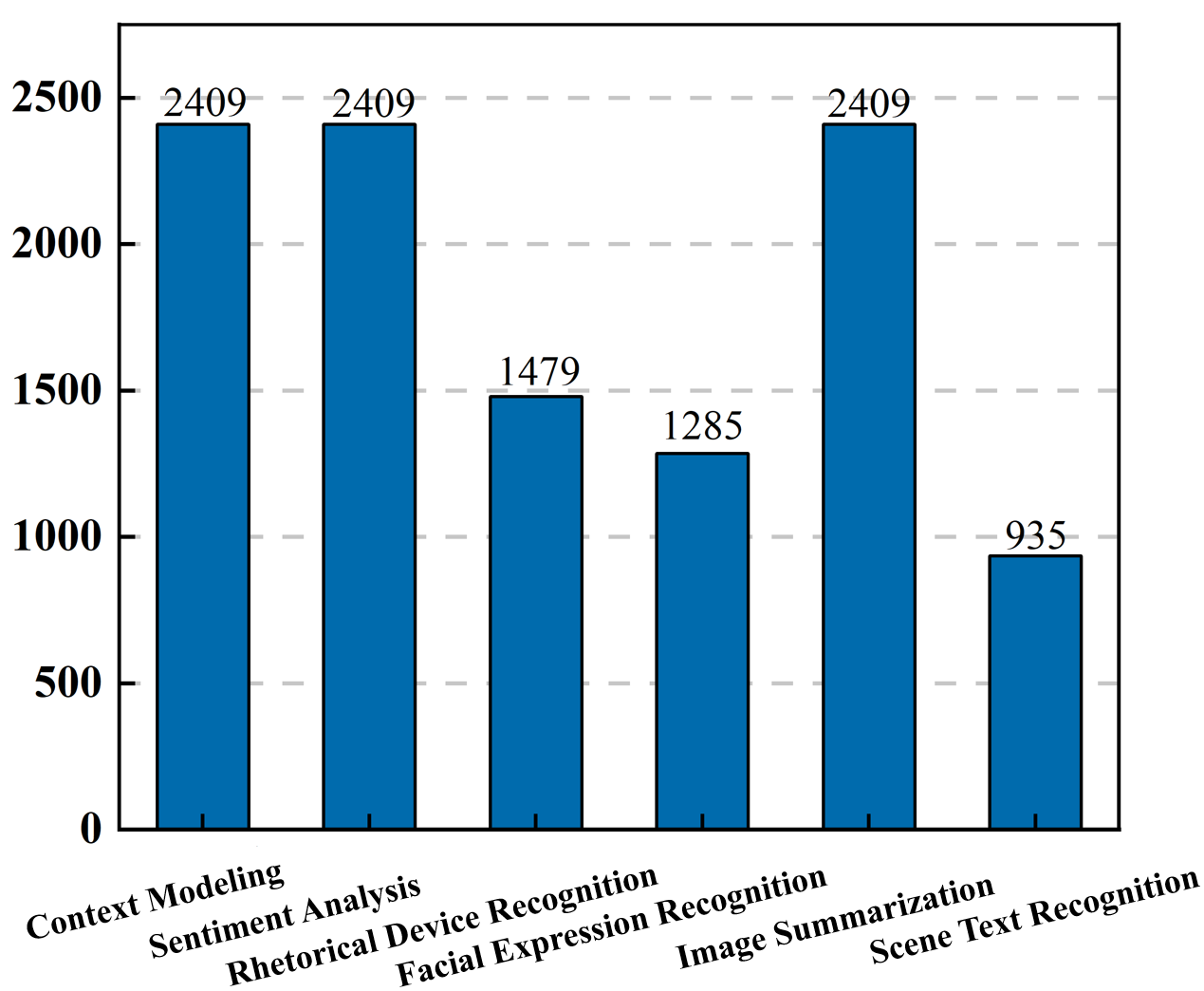}
    \caption{Agent call frequency on the MMSD 2.0 dataset.}
   \label{fig:Frequency}
\end{figure}

Fig.~\ref{fig:Frequency} shows the invocation frequency for each subtask agent. \textbf{Context Modeling}, \textbf{Sentiment Analysis}, and \textbf{Image Summarization} are triggered in all cases (2,409 times each), reflecting their necessity for every input. By contrast, \textbf{Rhetorical Device Recognition}, \textbf{Facial Expression Recognition}, and \textbf{Scene Text Recognition} are conditionally invoked, with call counts of 1,479, 1,285, and 935, respectively. The lower frequencies for the latter agents are due to the absence of rhetorical devices, facial expressions, or scene text in a significant proportion of samples.

This distribution highlights two important findings. First, the frequency of subtask invocation is closely linked to the nature of the input data (i.e., whether textual or visual clues are present). Second, agents such as Facial Expression Recognition and Scene Text Recognition—though not always activated—capture critical cues for sarcasm detection when relevant. These results further support the rationality of designing dedicated subtasks and specialized agents, as each contributes unique information to the overall multi-agent sarcasm understanding framework.

\subsection{Out-of-Distribution Results}
To evaluate the robustness and generalization of Commander-GPT, we conducted out-of-distribution sarcasm detection experiments on the SemEval 2018 Task 3 dataset. We selected two LLMs as the commander: the open-source MiniCPM-V2, and the SoTA model Claude-3.5 (serving as a new commander).

\begin{figure*}[!ht]
    \centering
    \subfloat
    {\includegraphics[width=2.7in]{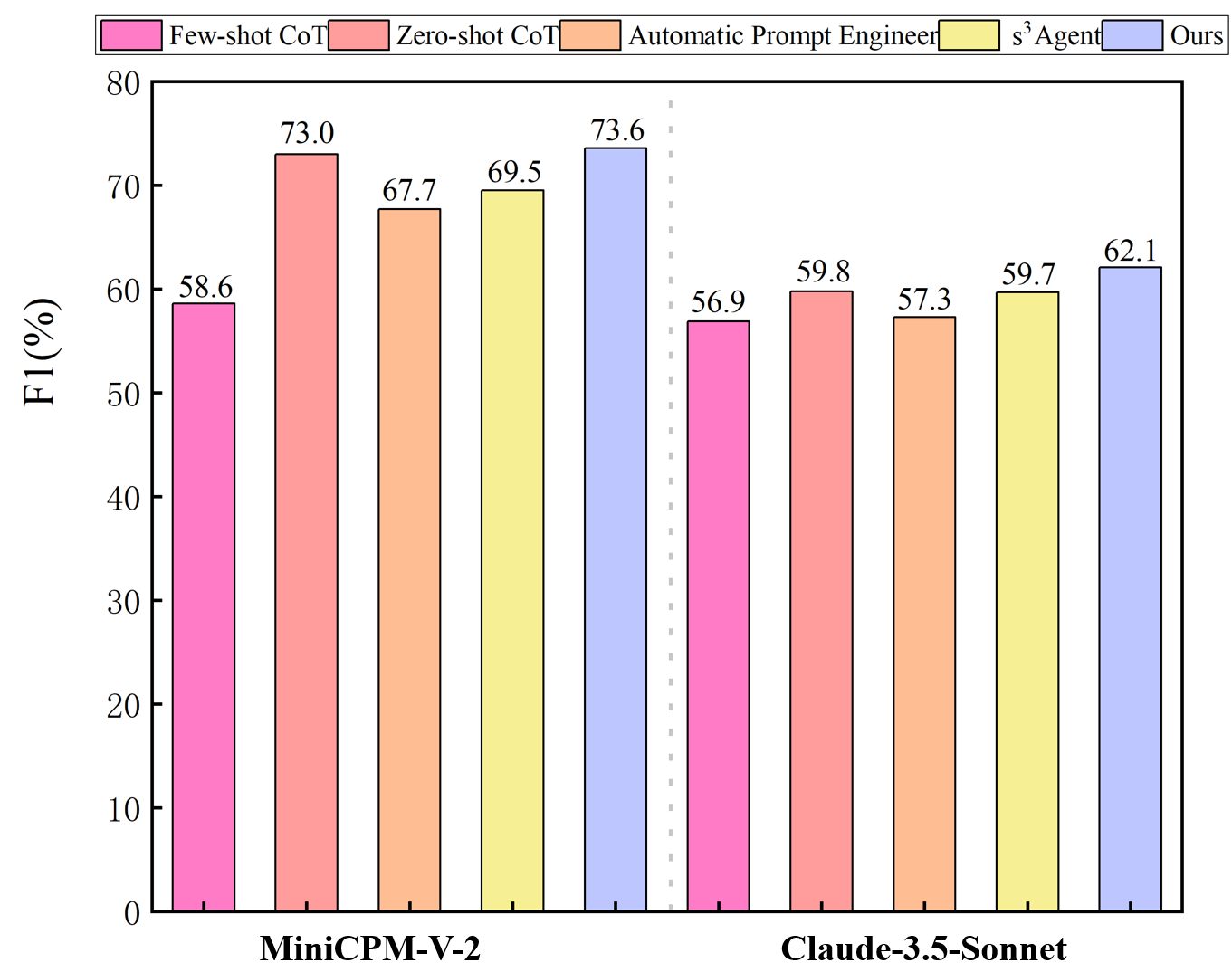}
    \label{fig: SinglePrompt}}
    \hfil 
    \subfloat{\includegraphics[width=2.7in]{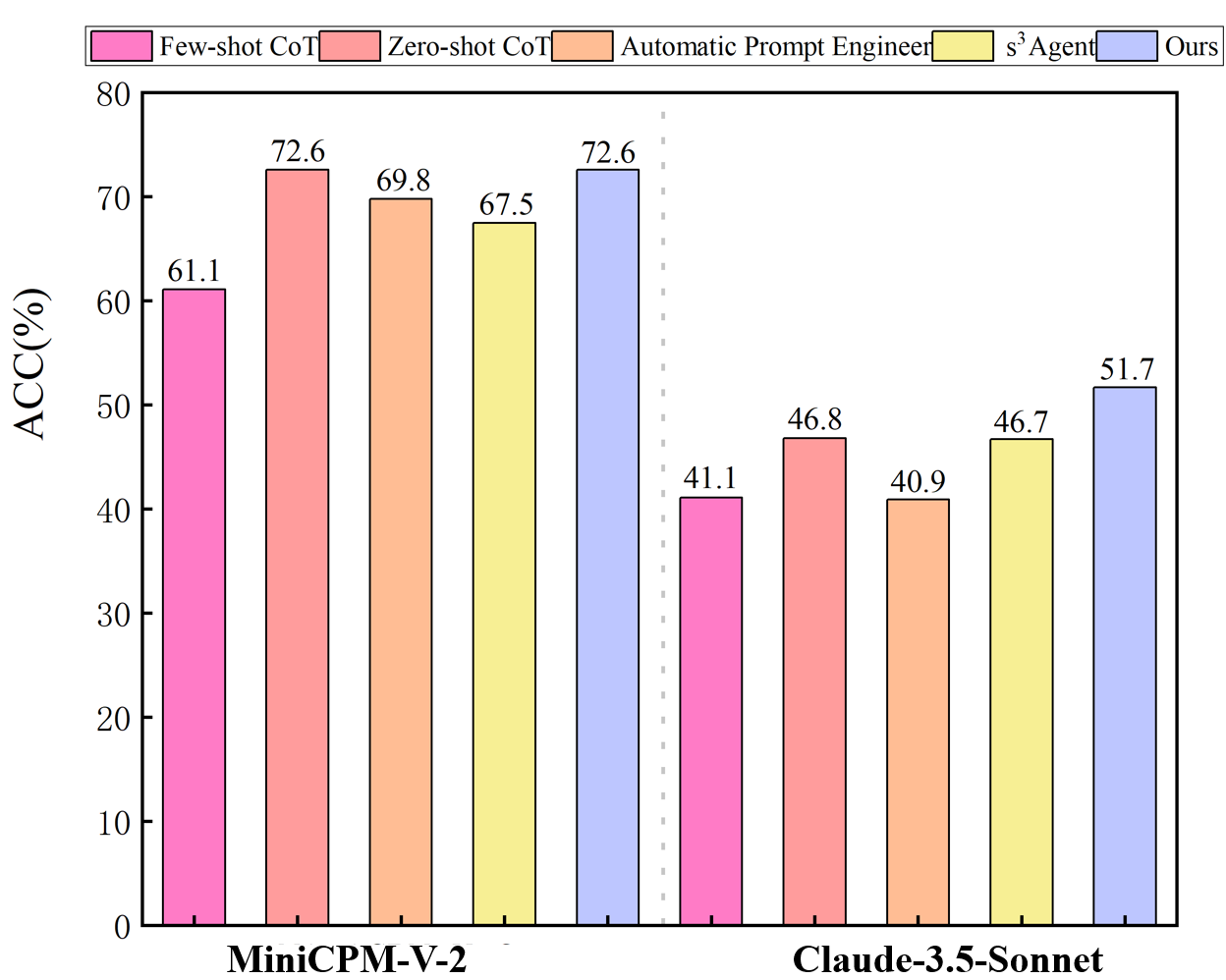}%
    \label{fig:MultiPrompt}}
    \caption{The experimental results of MiniCPM-V-2 and Claude-3 on SemEval 2018 Task 3 were statistically evaluated.}
    \label{fig:confusionmatrix2}
\end{figure*}

Fig.~\ref{fig:confusionmatrix2} reports the F1 and Acc. scores of all methods. On MiniCPM-V2, Commander-GPT achieves an F1 of 73.6\% and an Acc. of 72.6\%, outperforming all prompt-based and agent-based baselines. Specifically, compared to the strongest baseline (\textit{S}$^3$ Agent, F1 69.5\%, Acc. 67.5\%), Commander-GPT brings improvements of 4.1\% and 5.1\%, respectively.

When using Claude 3.5 as the commander, Commander-GPT still maintains the leading performance, achieving 62.1\% F1 and 51.7\% Acc. This exceeds the best baseline (Zero-shot CoT, F1 59.8\%, Acc. 46.8\%) by 2.3\% and 4.9\%, respectively. It is worth noting that the performance boost remains consistent even when switching to a more powerful closed-source model as the commander, demonstrating the flexibility and model-agnostic nature of our framework.

Overall, these results confirm that Commander-GPT can robustly generalize to new domains and leverage the advantages of different backbone LLMs. Whether with open-source or SOTA commercial models as the commander, our framework consistently delivers state-of-the-art results in sarcasm detection.

\subsection{The Impact of The Number of Subtasks}
\begin{figure*}[!ht]
    \centering
    \subfloat
    {\includegraphics[width=2.7in]{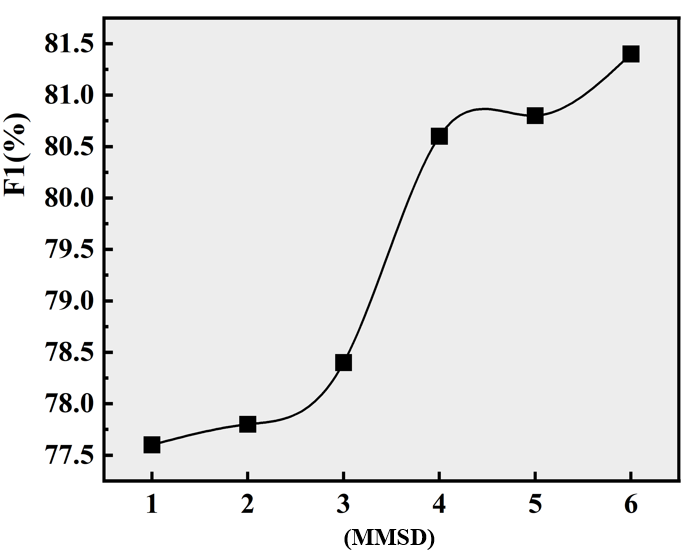}
    \label{fig: SinglePrompt}}
    \hfil 
    \subfloat{\includegraphics[width=2.7in]{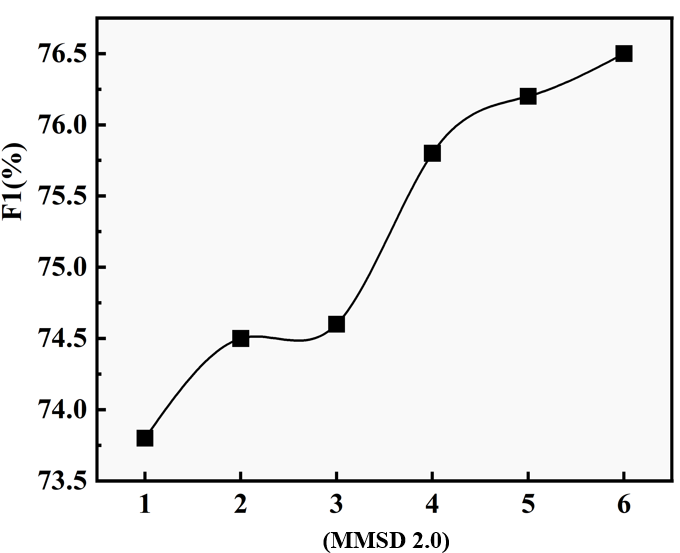}%
    \label{fig:MultiPrompt}}
    \caption{F1 score variation curves for the MMSD and MMSD 2.0 datasets with varying numbers of submodules, where the modules are added in the following order: Context Modeling, Sentiment Analysis, Image Summarization, Scene Text Recognition, Rhetorical Device Recognition, and Facial Expression Recognition.
}
    \label{fig:number}
\end{figure*}

From the experimental results in Fig.~\ref{fig:number}, it can be seen that as the number of submodules increases, the model’s F1 score on the MMSD and MMSD 2.0 datasets follows a three‐stage trend of “slow start, rapid rise, then gentle plateau with a final slight boost.” The first two basic modules—Context Modeling and Sentiment Analysis—bring only limited gains until the introduction of Image Summarization and Scene Text Recognition triggers a substantial jump, with Scene Text Recognition especially filling the gap between textual and visual information. Adding Rhetorical Device Recognition and Facial Expression Recognition thereafter still yields improvements, but with diminishing marginal returns. On MMSD 2.0, thanks to higher data quality, the modules cooperate more smoothly and deliver steadier performance gains. Therefore, under resource constraints, priority should be given to Image Summarization and Scene Text Recognition, followed by Rhetorical Device Recognition and Facial Expression Recognition, while cross‐modal fusion strategies can be explored to further unlock the potential of later modules.

\subsection{Routing Score Visualization}

\begin{figure*}[!ht]
    \centering
    \includegraphics[width=4.8in]{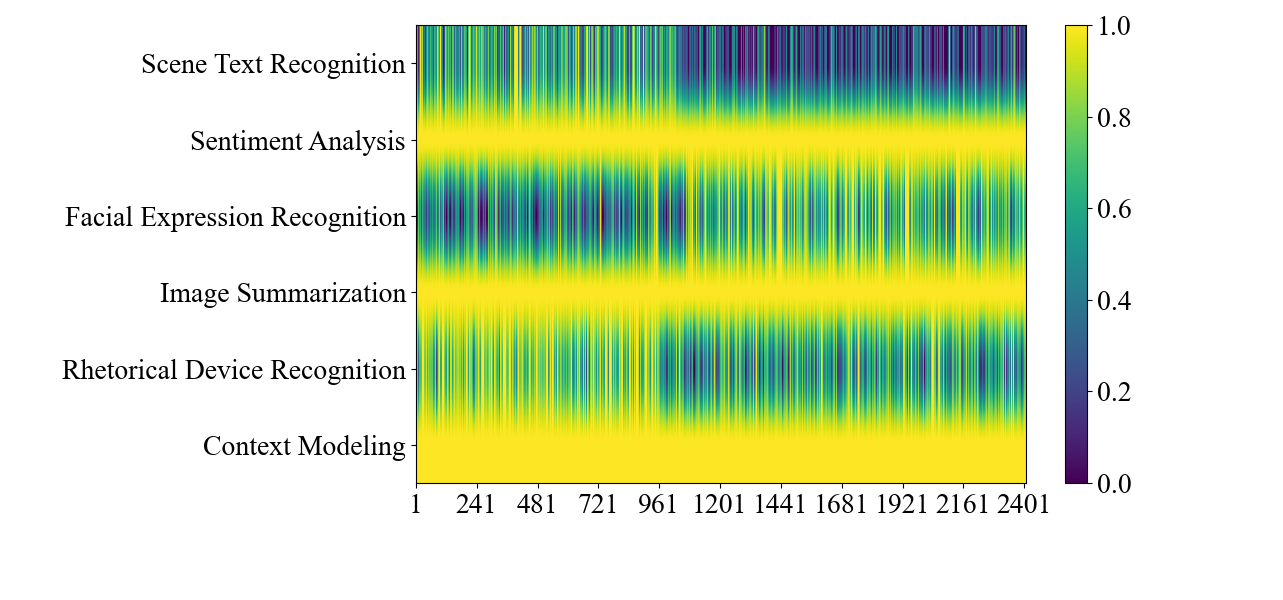}
    \caption{Heatmap of subtask agent counts on the MMSD 2.0 dataset.}
   \label{fig:heatmap}
\end{figure*}

To further interpret the decision mechanism of Commander-GPT, we visualize the routing scores assigned to each subtask agent for representative samples from the MMSD 2.0 dataset. Fig.~\ref{fig:heatmap} presents a heatmap where each column corresponds to an input sample and each row corresponds to one of the six subtask agents. The color intensity indicates the routing score (i.e., the normalized weight assigned by the router to each agent).

The heatmap reveals clear patterns in the agent assignment. For samples containing explicit rhetorical devices, the router assigns high scores to the \textit{Rhetorical Device Recognition} agent. Samples with prominent facial expressions or scene text result in higher weights for the \textit{Facial Expression Recognition} and \textit{Scene Text Recognition} agents, respectively. In contrast, samples lacking such cues rely more on the \textit{Context Modeling} and \textit{Sentiment Analysis} agents. This dynamic and adaptive score distribution demonstrates that the router effectively identifies and leverages the most relevant subtasks for each input, thereby improving interpretability and robustness.

Overall, the routing score visualization provides direct evidence that the proposed multi-agent architecture can selectively integrate different information sources based on the input characteristics, supporting both the transparency and effectiveness of our approach.

\subsection{Case Study}
Table~\ref{tab:casestudy1} and Table~\ref{tab:casestudy2} presents examples of sarcasm detection predictions across multiple LLM-based methods. We analyze the key error patterns in model performance below.

\textbf{(1) Contextual misunderstanding.}  
Models such as Zero-shot CoT struggle with sarcasm that relies on implicit contextual cues. In Example 2 (``the pa welcome center is hopping today''), the word ``hopping'' sarcastically contrasts with the likely quiet reality of a welcome center; however, Zero-shot CoT misclassified this instance as non-sarcastic. Similarly, Example 6 (``a bit of a \#mantub revival'') hinges on a niche cultural reference that AutoPE failed to interpret, resulting in an incorrect prediction.

\textbf{(2) Literal interpretation.}  
Models frequently prioritize surface-level semantics over implied tone. For instance, Example 5 (``do you suffer from this related problem? :)\,'') uses both an emoji and phrasing to mock faux concern, but Plan-and-Solve misclassified it as non-sarcastic. Likewise, Example 9 (``protein pizza coming soon!'') uses hyperbolic language to mock food trends, yet Golden labeled it as non-sarcastic, reflecting an over-reliance on literal keyword associations.

\textbf{(3) Bias toward hashtag patterns.}  
Some models, such as GenKPrompt, exhibit a bias toward treating hashtags as sarcasm markers. Example 1 (``\#nihilistmemes'') was correctly classified by Golden but mislabeled by S\textsuperscript{2} Agent due to overemphasis on the hashtag. Conversely, Example 8 (``\#womensmarch'') combines serious activism with sarcastic phrasing (``welcome to your first day''), leading our model to misinterpret it as non-sarcastic, likely because the model conflated the topic-specific hashtag with genuine intent.

This analysis highlights the need for LLMs to better disentangle contextual subtleties, mitigate literal bias, and reduce overfitting to surface-level markers such as hashtags in sarcasm detection tasks.

\begin{table*}[t]
 \captionof{table}{Typical examples for case study.} 
\label{tab:casestudy1}
\centering
\scalebox{0.72}{
\begin{tabular}{c p{4.5cm} m{3.6cm} c c c c}
\hline
\textbf{ID} & \textbf{Text} & \textbf{Image} & \textbf{Golden} & \textbf{Zero-shot COT} & \textbf{AutoPE} & \textbf{Plan-and-Solve} \\
\hline
1       & happy new year , everyone ! xoxo \# year2016 \# partyhard \# nihilistmemes            & \includegraphics[width=3cm]{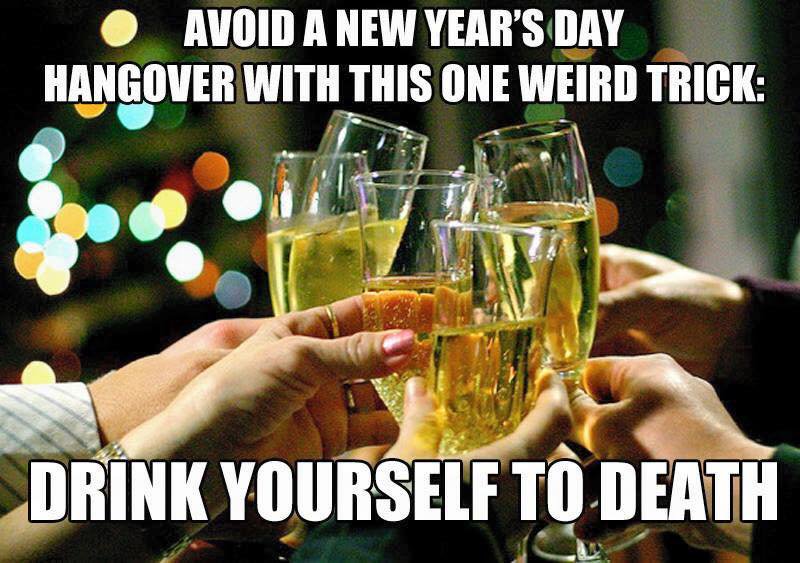}     & Sarcastic     & \multicolumn{1}{c}{\faCheckSquare}                   & \multicolumn{1}{c}{\faCheckSquare}                            & \multicolumn{1}{c}{\faCheckSquare}    \\
2       & the pa welcome center is hopping today .                                              & \includegraphics[width=3cm]{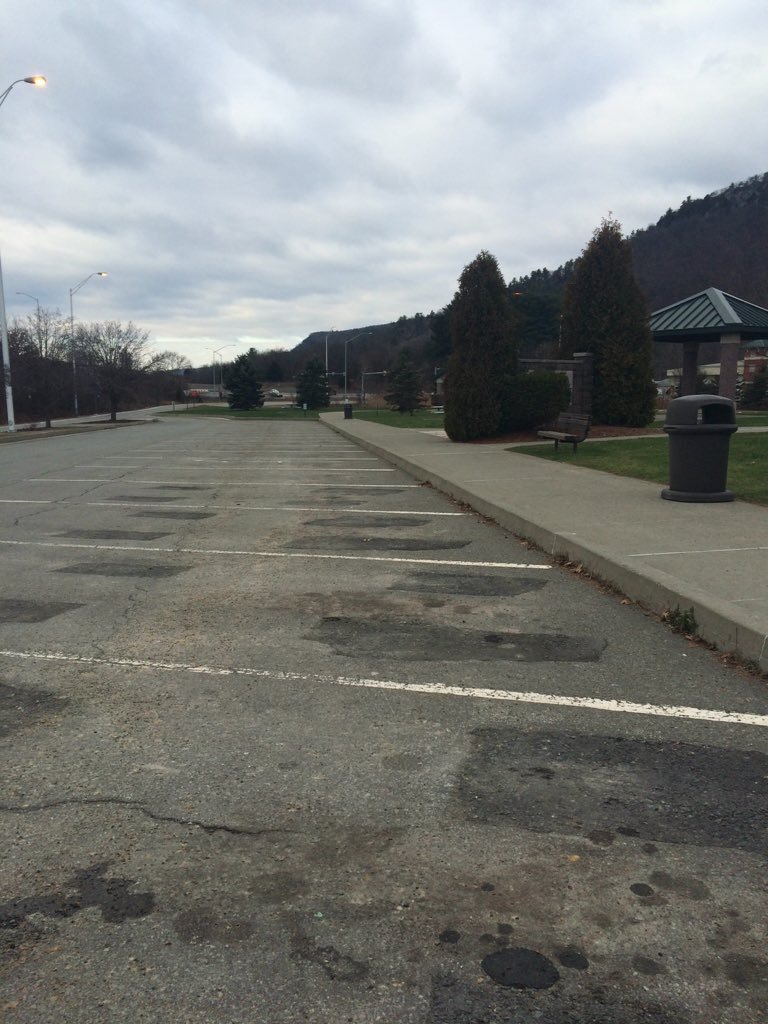}     & Sarcastic     & \multicolumn{1}{c}{\faTimes}                         & \multicolumn{1}{c}{\faTimes}                                  & \multicolumn{1}{c}{\faCheckSquare}    \\
3       & because every mildly successful cgi film needs an animated spinoff .                  & \includegraphics[width=3cm]{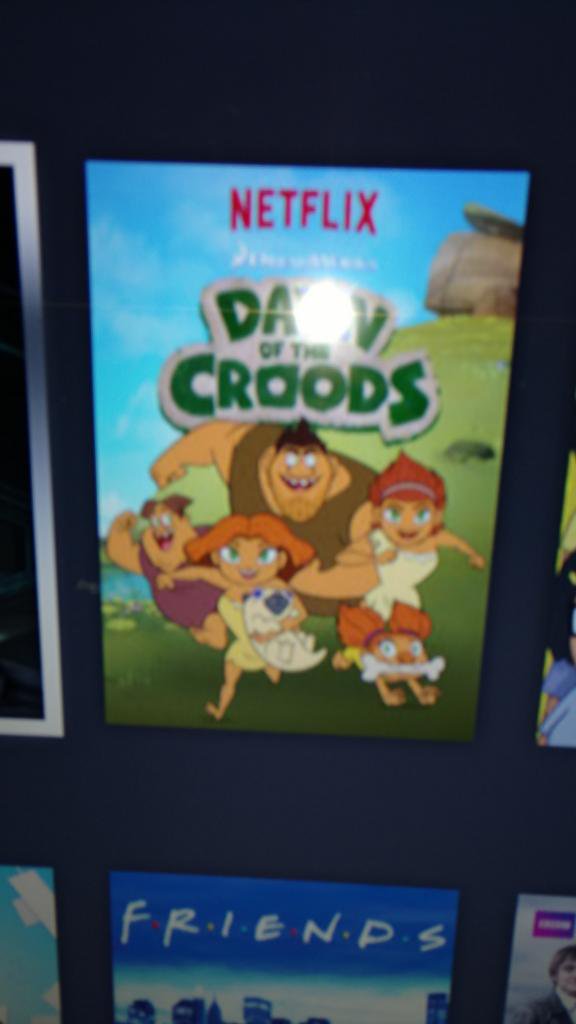}     & Sarcastic     & \multicolumn{1}{c}{\faCheckSquare}                   & \multicolumn{1}{c}{\faCheckSquare}                            & \multicolumn{1}{c}{\faCheckSquare}    \\
4       & dad 's handy work . can 't tell at all .                                              & \includegraphics[width=3cm]{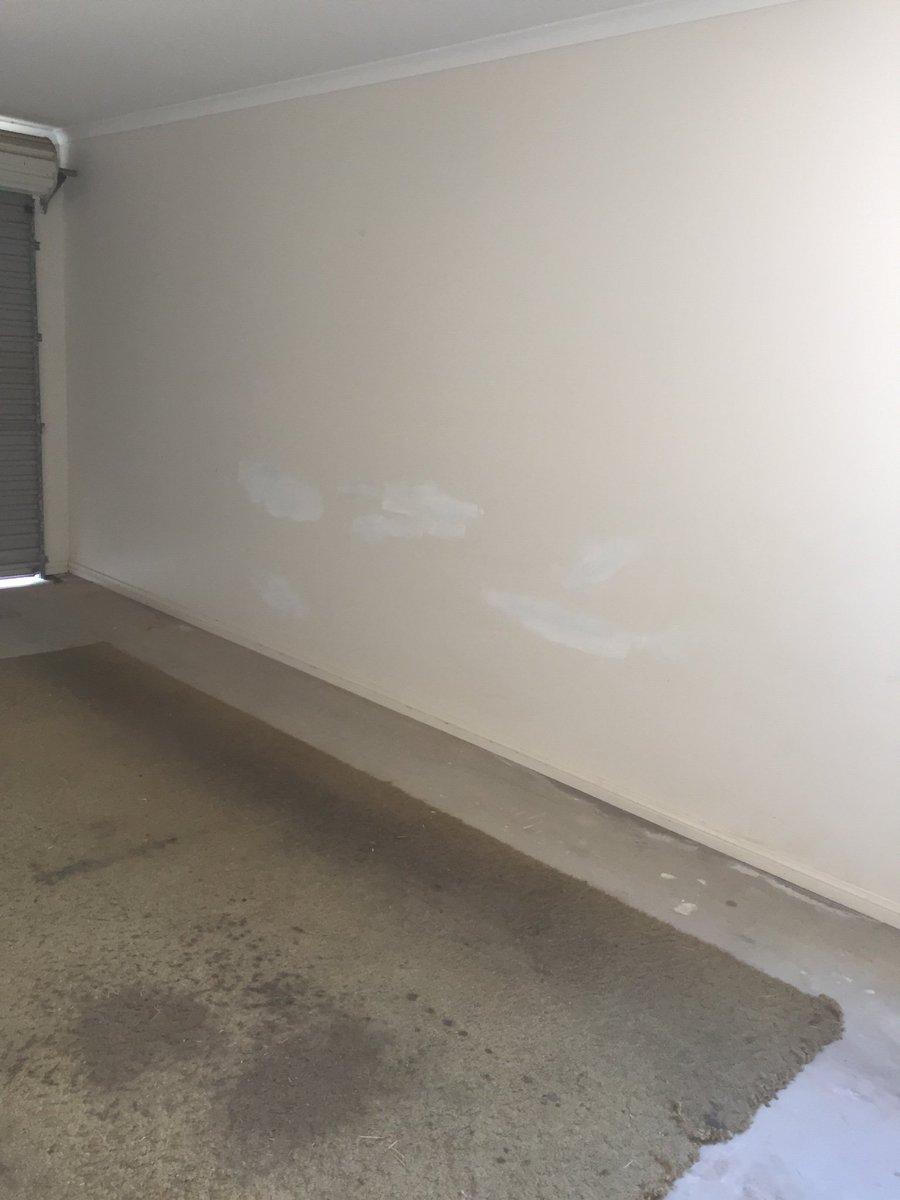}     & Sarcastic     & \multicolumn{1}{c}{\faCheckSquare}                   & \multicolumn{1}{c}{\faCheckSquare}                            & \multicolumn{1}{c}{\faTimes}          \\
5       & do you suffer from this related problem ? :)                                          & \includegraphics[width=3cm]{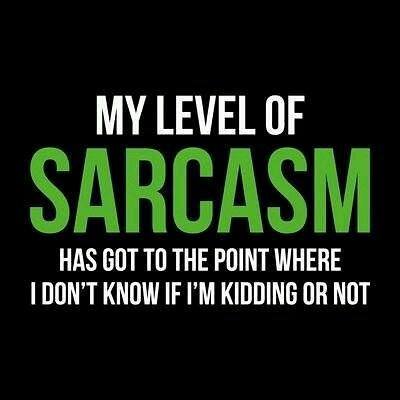}     & Sarcastic     & \multicolumn{1}{c}{\faCheckSquare}                   & \multicolumn{1}{c}{\faCheckSquare}                            & \multicolumn{1}{c}{\faCheckSquare}    \\
6       & meanwhile , back in the \# cbb house , there 's been a bit of a \# mantub revival ... & \includegraphics[width=3cm]{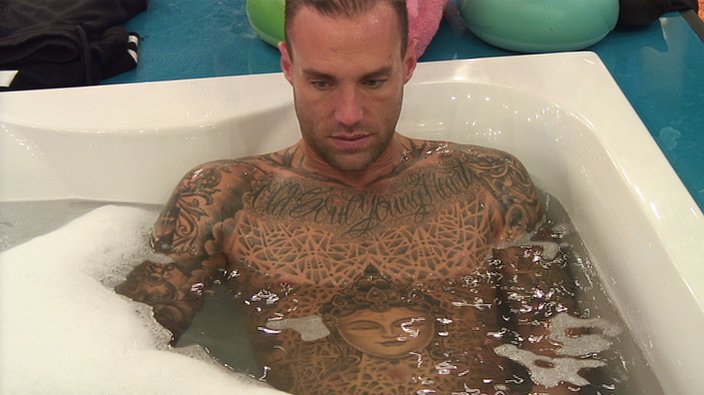}     & Non-sarcastic & \multicolumn{1}{c}{\faCheckSquare}                         & \multicolumn{1}{c}{\faCheckSquare}                                  & \multicolumn{1}{c}{\faTimes}          \\
7       & harry houdini 1914 . colorized by dana r keller .                                     & \includegraphics[width=3cm]{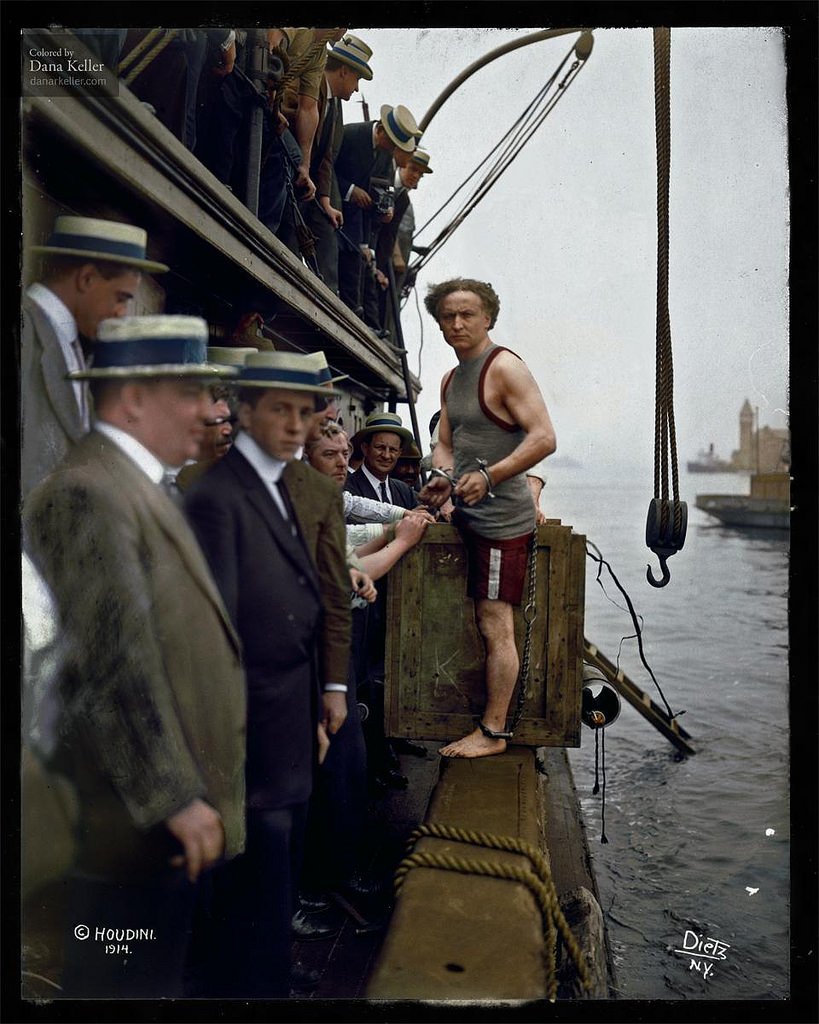}     & Non-sarcastic & \multicolumn{1}{c}{\faCheckSquare}                         & \multicolumn{1}{c}{\faCheckSquare}                                  & \multicolumn{1}{c}{\faCheckSquare}          \\\hline
\end{tabular}
}
\end{table*}

\begin{table*}[t]
\caption{Typical examples for case study.} 
\label{tab:casestudy2}
\centering
\scalebox{0.72}{
\begin{tabular}{c p{4.5cm} m{3.6cm} c c c c}
\hline
\textbf{ID} & \multicolumn{1}{c}{\textbf{Text}}                                                                                 & \textbf{Image} & \textbf{Golden}        & \textbf{GenKPrompt} & \textbf{$S^3$ Agent} & \multicolumn{1}{c}{\textbf{Ours}}           \\ \hline
1       & happy new year , everyone ! xoxo \# year2016 \# partyhard \# nihilistmemes            & \includegraphics[width=3cm]{casestudy/682716753374351360.jpg}     & Sarcastic     & \multicolumn{1}{c}{\faCheckSquare}                   & \multicolumn{1}{c}{\faCheckSquare}                            & \multicolumn{1}{c}{\faCheckSquare}    \\
2       & the pa welcome center is hopping today .                                              & \includegraphics[width=3cm]{casestudy/683295117847969793.jpg}       & Sarcastic     & \multicolumn{1}{c}{\faCheckSquare}                   & \multicolumn{1}{c}{\faTimes}                                  & \multicolumn{1}{c}{\faCheckSquare}    \\
3       & because every mildly successful cgi film needs an animated spinoff .                  & \includegraphics[width=3cm]{casestudy/684160451572281344.jpg}       & Sarcastic     & \multicolumn{1}{c}{\faCheckSquare}                   & \multicolumn{1}{c}{\faCheckSquare}                            & \multicolumn{1}{c}{\faCheckSquare}    \\
4       & dad 's handy work . can 't tell at all .                                              & \includegraphics[width=3cm]{casestudy/684633048483106816.jpg}     & Sarcastic     & \multicolumn{1}{c}{\faCheckSquare}                   & \multicolumn{1}{c}{\faTimes}                                  & \multicolumn{1}{c}{\faCheckSquare}    \\
5       & do you suffer from this related problem ? :)                                          & \includegraphics[width=3cm]{casestudy/685601583879598084.jpg}     & Sarcastic     & \multicolumn{1}{c}{\faCheckSquare}                   & \multicolumn{1}{c}{\faCheckSquare}                            & \multicolumn{1}{c}{\faCheckSquare}    \\
6       & meanwhile , back in the \# cbb house , there 's been a bit of a \# mantub revival ... & \includegraphics[width=3cm]{casestudy/822950612908343298.jpg}     & Non-sarcastic & \multicolumn{1}{c}{\faCheckSquare}                         & \multicolumn{1}{c}{\faCheckSquare}                                  & \multicolumn{1}{c}{\faCheckSquare}          \\
7       & harry houdini 1914 . colorized by dana r keller .                                     & \includegraphics[width=3cm]{casestudy/822950757678845952.jpg}     & Non-sarcastic & \multicolumn{1}{c}{\faCheckSquare}                         & \multicolumn{1}{c}{\faCheckSquare}                                  & \multicolumn{1}{c}{\faCheckSquare}          \\\hline
\end{tabular}
}
\end{table*}

\section{Conclusion}
In this paper, we proposed Commander-GPT, a novel multi-agent routing framework for multimodal sarcasm detection. Our approach decomposes the complex sarcasm detection task into cognitively meaningful subtasks, dynamically routing each input to the most suitable specialist agent. The central commander integrates information from all subtask agents, enabling adaptive and fine-grained reasoning across both textual and visual modalities. Extensive experiments on multiple benchmarks demonstrate that Commander-GPT consistently achieves state-of-the-art performance. Further analyses confirm the effectiveness, robustness, and strong generalization ability of our framework. We believe this work offers new insights into explainable and modular design for sarcasm and sentiment analysis.

\textbf{Limitations.}
Despite its advantages, Commander-GPT relies on the quality and diversity of available subtask agents; limited agent expressiveness may restrict performance in highly novel or ambiguous scenarios. In addition, the current routing strategy depends on labeled subtasks and may incur increased computational cost compared to end-to-end monolithic models. Future work could explore automatic subtask discovery, more efficient routing mechanisms.

\section*{Acknowledgment}
This paper is partly supported by a grant under Hong
Kong RGC Theme-based Research Scheme (project no. T45-401/22-N). This work is supported by National Science Foundation of China under grant No. 62006212, Fellowship from the China Postdoctoral Science Foundation (2023M733907), Natural Science Foundation of Hunan Province of China (242300421412), Foundation of Key Laboratory of Dependable Service Computing in Cyber-Physical-Society (Ministry of Education), Chongqing University (PJ.No: CPSDSC202103).

\bibliography{colm2025_conference}
\bibliographystyle{colm2025_conference}

\end{document}